\documentclass{article}

\usepackage[preprint]{neurips_2025}

\usepackage[utf8]{inputenc} 
\usepackage[T1]{fontenc}    
\usepackage{hyperref}       
\usepackage{url}            
\usepackage{booktabs}       
\usepackage{amsfonts}       
\usepackage{natbib}         
\usepackage{nicefrac}       
\usepackage{microtype}      
\usepackage{xcolor}         

\usepackage{amsmath}
\usepackage{amsthm}
\usepackage{bm}

\newtheorem{theorem}{Theorem}
\usepackage{pifont}
\usepackage{amssymb}
\usepackage{graphicx}
\usepackage{subfig}
\usepackage{algorithm}
\usepackage{multirow}
\usepackage{color}
\usepackage{enumitem}

\usepackage{algpseudocode}
\usepackage{tabularx}       
\usepackage{caption}          
\usepackage{float}            
\usepackage{mdframed}         
\usepackage{adjustbox}        

\title{SchoenbAt: Rethinking Attention with\\ Polynomial basis}

\mdfdefinestyle{MyFrame}{%
    linecolor=black,
    linewidth=1pt,
    leftmargin=0pt,
    innerleftmargin=10pt,
    innerrightmargin=10pt,
    skipabove=10pt,
    skipbelow=10pt
}
\author{
Yuhan Guo$^{1,2}$, Lizhong Ding$^{1,3}$, Yuwan Yang$^1$, Xuewei Guo$^1$\\
$^1$Beijing Institute of Technology\\
$^2$3220231188@bit.edu.cn,
$^3$lizhong.ding@outlook.com
}

\begin{document}
\bibliographystyle{unsrt}

\maketitle

\begin{abstract}
Kernelized attention extends the attention mechanism by modeling sequence correlations through kernel functions, making significant progresses in optimizing attention. Under the guarantee of harmonic analysis theory, kernel functions can be expanded with basis functions, inspiring random feature-based approaches to enhance the efficiency of kernelized attention while maintaining predictive performance. However, current random feature-based works are limited to the Fourier basis expansions under Bochner's theorem. We propose Schoenberg's theorem-based attention (SchoenbAt), which approximates dot-product kernelized attention with the polynomial basis under Schoenberg's theorem via random Maclaurin features and applies a two-stage regularization to constrain the input space and restore the output scale, acting as a drop-in replacement of dot-product kernelized attention. Our theoretical proof of the unbiasedness and concentration error bound of SchoenbAt supports its efficiency and accuracy as a kernelized attention approximation, which is also empirically validated under various random feature dimensions. Evaluations on real-world datasets demonstrate that SchoenbAt significantly enhances computational speed while preserving competitive performance in terms of precision, outperforming several efficient attention methods.
\end{abstract}

\section{Introduction}

In recent years, attention mechanisms \cite{attention} have made significant advancements in several areas of machine learning including natural language processing \cite{nlp1,nlp2,nlp3}, computer vision \cite{cv4,cv5,cv6,cv7} and bioinformatics \cite{bio1,bio2,bio3}, particularly with the success of the Transformer architecture \cite{transformer} in Large-Language Models \cite{llm1,llm2,llm3,llm4}. Kernelized attention \cite{trnn,dissection} is a widely studied variant of attention \cite{linformer,Nyströmformer,skyformer}, which generalizes the function capturing sequence dependencies to a kernel function, thereby offering new perspectives for broadening the horizons of research \cite{ka1,ka2,ka3}. 

The expansion of the kernel function is one of the key techniques for efficient kernelized attention. As a classical theoretical result in harmonic analysis, Bochner's theorem \cite{bochner} guarantees that a \textit{shift-invariant kernel} is an expansion with Fourier bases, motivating random Fourier feature methods \cite{randomFeature1,randomFeature2,randomFeature3,randomFeature4,randomFeature5,gausappro} which approximate the kernel function through nonlinear random mappings, significantly improving computational efficiency while maintaining accuracy \cite{rfa,performer,rfapp1,rfapp2,rfapp3,spectraformer}. However, the attention mechanism inherently models sequence dependencies through a \textit{dot-product kernel}, and works based on Fourier expansions require additional computation to transform the dot-product kernel into a shift-invariant kernel, limiting their practical utility.

We observe that \textit{the polynomial basis expansion of dot-product kernelized attention} can be theoretically guaranteed by Schoenberg’s theorem \cite{schoenberg}, which motivates our proposed \underline{Schoen}berg’s theorem-\underline{b}ased \underline{At}tention mechanism (\textbf{SchoenbAt}). Starting from Mercer’s theorem \cite{mercer}, we reinterpret kernelized attention through the lens of basis function expansions, establishing a unified theoretical framework. Harmonic analysis provides that kernel functions can be expressed as expansions over specific basis functions, with Bochner’s theorem and Schoenberg’s theorem yielding Fourier and polynomial basis representations, respectively. In this work, we introduce a novel polynomial basis expansion for kernelized attention. Building on the unbiased approximation of dot-product kernels using Random Maclaurin Features (\textbf{RMF}) \cite{rmf}, we design Random Maclaurin Feature Attention (\textbf{RMFA}) as the core component of SchoenbAt, which reduces the computational complexity of kernelized attention by restructuring its computational graph and fully leveraging computational parallelism. Since Schoenberg’s theorem requires the input space of RMFA to be bounded, SchoenbAt incorporates a Pre-Post Scaling Batch Normalization (\textbf{ppSBN}) mechanism, inspired by Batch Normalization \cite{batchNorm}. This mechanism scales the input with trainable parameters to satisfy the theoretical constraints of Schoenberg’s theorem, while also restoring the original output magnitude, ensuring that SchoenbAt remains an unbiased estimator of the kernelized attention when the input space is unconstrained. We provide a rigorous proof of the unbiasedness and establish an approximation error bound as a function of the random feature dimension, laying a solid theoretical foundation for SchoenbAt. Our numerical experiments report the approximation errors of SchoenbAt with five different kernel functions across varying random feature dimensions, as well as the speedup ratios on simulated data of different lengths. To evaluate the performance of SchoenbAt on real-world datasets, we applied SchoenbAt to a basic Transformer \cite{skyformer} and compared it with various effective attention methods on the Long Range Arena (LRA) benchmark \cite{lra}, which includes five tasks covering both natural language processing (Text \cite{lratext}, ListOps \cite{listops}, Retrieval \cite{lraretrieval}) and computer vision (Pathfinder \cite{lrapathfinder}, Image \cite{lraimage}). Our results show that SchoenbAt delivers substantial acceleration and maintains competitive accuracy across diverse domains.

\section{Background}

\textbf{Notation:}
In this paper, we denote matrices in uppercase bold letters like $\bm{X}$ and vectors in lowercase letters like $x$. Unless specifically mentioned, our vectors \(x\in \mathbb{R}^d \), and the input to the attention mechanism, namely the query, key, and value matrices, is denoted as $\bm{Q},\bm{K},\bm{V}\in \mathbb{R}^{n\times d}$. We discuss a kernel function \( \mathcal{K}: \mathbb{R}^d \times \mathbb{R}^d \to \mathbb{R} \) in its corresponding reproducing kernel Hilbert space \( \mathcal{H} \). A function defined on a vector space $\mathbb{R}^d$ can also accept tensor inputs of arbitrary dimensions, acting on the last dimension. Let $\ell_\rho (0,\xi)$ denotes the set $\{ {\left \| \bm{X} \right \|}_\rho \le \xi \}$. 

\subsection{Kernelized Attention}

Derived from the widely used and fundamental softmax attention \cite{attention}, which models the correlation between $\bm{Q}$ and $\bm{K}$ with $\exp(\cdot)$, constructing a convex combination of \( \bm{V} \) as the output:
\begin{equation*}
{\rm attn}_{\rm Softmax}(\bm{Q},\bm{K},\bm{V})={ \sum_{i=1}^{n}\frac{\exp(\bm{Q} \bm{K}_i^\top/\sqrt{d}) \bm{V}_i }{ \sum_{j=1}^{n}\exp(\bm{Q} \bm{K}_j^\top/\sqrt{d}) \mathbf{1}_d^\top} },
\end{equation*}
where $\mathbf{1}_d=[1,1,...,1]^\top\in\mathbb{R}^d$ and $\frac{\bm{X}}{\bm{Y}}$ means element-wise division, kernelized attention \cite{trnn,dissection} proposes to generalize the function modeling sequence similarity into a kernel function, resulting in a linear combination of \( \bm{V} \) weighted by the normalized \( \mathcal{K}(\bm{Q}, \bm{K}) \):
\begin{equation}
{\rm attn}_{\mathcal{K}}(\bm{Q},\bm{K},\bm{V})={ \sum_{i=1}^{n}\frac{\mathcal{K}(\bm{Q} \bm{K}_i^\top/\sqrt{d}) \bm{V}_i }{ \sum_{j=1}^{n}\mathcal{K}(\bm{Q} \bm{K}_j^\top/\sqrt{d})\mathbf{1}_d^\top } }.
\nonumber
\end{equation}

When \( \mathcal{K}(\cdot)=\exp(\cdot) \), kernelized attention degenerates into softmax attention. Kernelized attention transforms the computation of attention into kernel functions, offering another perspective for studying attention. For example, efficient attention based on Nyström method views the attention matrices as kernel matrices and approximates them by low-rank approximations \cite{rel1,informer,Nyströmformer,skyformer}.

\begin{figure}[t]
\centering
\includegraphics[width=0.98\textwidth]{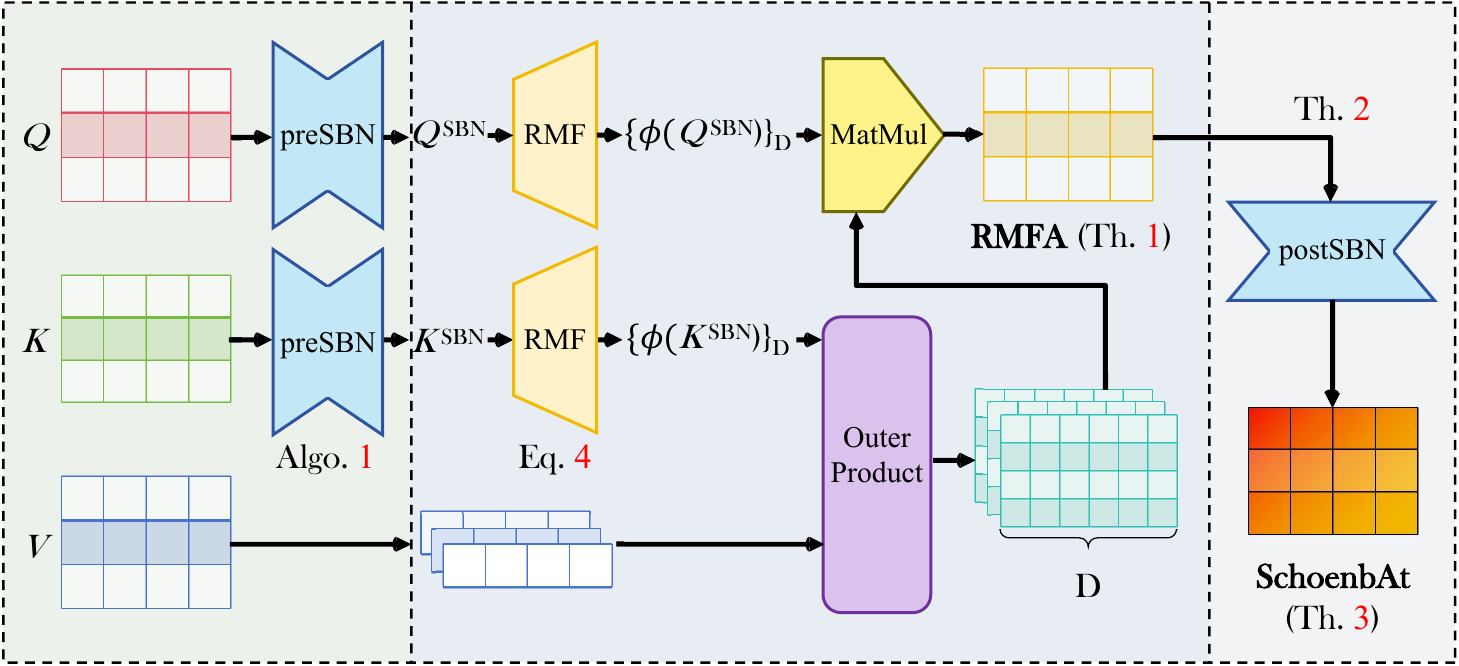}
\caption{The module design structure of SchoenbAt.}
\label{schnepic}
\end{figure}

\subsection{Kernel Expansions in Fourier Basis}

We study kernelized attention from the perspective of basis functions \cite{basis1,basis2,basis3,gausbasis}. As a classical result in harmonic analysis, Bochner's theorem \cite{bochner} ensures that shift-invariant kernels can be expanded as a Fourier transform of a non-negative measure $m:\mathbb{R}^d \to \mathbb{R}_{\ge 0}$:
\begin{equation}
\label{shift-inv}
\mathcal{K}(x-y)=\int_{\mathbb{R}^d}m(\omega )e^{i\omega^\top (x-y)}\mathrm{d}\omega, m(\cdot)\ge 0. 
\end{equation}

Equation (\ref{shift-inv}) allows one to represent a shift-invariant kernel as an expansion with Fourier bases, where the measure \( m(\cdot) \) can be seen as the weight for each Fourier basis. Rahimi (2007) proposed a random feature sampling method \cite[Algorithm 1]{randomFeature1} that unbiasedly estimates the Fourier transform when the weights are probability measures, thus providing an approximation of a shift-invariant kernel.

By properly scaling the input and representing the exponential function by a Gaussian kernel as
\begin{align}
\nonumber
\begin{split}
\exp(\bm{Q}\bm{K}^\top\big/\sqrt{d})
&=\exp\!\left (\! \frac{\bm{Q}^\top \bm{Q}+\bm{K}^\top \bm{K}}{2\sqrt{d} }\! +\!\bm{C} \!\right )\! \exp\!\left (\! -\frac{\left \| \bm{Q}-\bm{K} \right \|^2 }{2\sqrt{d} } \! \right ) \\
&=\exp\left [ \left ( {\bm{Q}^\top \bm{Q}+\bm{K}^\top \bm{K}} \right ) /{2\sqrt{d} } +\bm{C} \right ]\mathcal{K}(\bm{Q}-\bm{K}),
\end{split}
\end{align}

\noindent where $\bm{C}\in\{\bm{X}|\bm{X}\in\mathbb{R}^{d\times d}\wedge{\rm tr}(\bm{X})=0\}$, a series of previous works \cite{rfa,performer,spectraformer} have optimized kernelized attention based on Bochner's theorem at the cost of introducing additional computations.

\section{SchoenbAt}

In this section, we first attempt to understand the expansion of kernel functions through a general expression that applies to both continuous and discrete domains, where Bochner's theorem and Schoenberg's theorem serve as specific instances. Based on the polynomial basis expansion of dot-product kernelized attention, we design SchoenbAt which employs the RMFA and ppSBN mechanisms to improve computational efficiency while maintaining performance. We derive theoretical results analyzing the unbiasedness and approximation error of SchoenbAt to guarantee its validity.

\subsection{Harmonic Expansions of SchoenbAt}

The kernel function of kernelized attention admits a general expansion form. Let \(\lambda\) be an eigenvalue of $\mathcal{K}(\cdot)$ and \(\psi(\cdot)\) be the corresponding eigenfunction, according to Mercer's theorem \cite{mercer}, $\{ \sqrt{\lambda_i}\psi_i \}_{i=1}^\infty$ form an orthogonal function basis for the Hilbert space $\mathcal{H}$.

In most cases, it is difficult to describe \(\psi(\cdot)\) explicitly \cite{gausbasis}. A harmonic analysis theorem ensures an expansion of a kernel function as a weighted combination of basis functions, thereby driving the research and development of kernelized attention under its theoretical foundation. 

We propose that, for a given domain \(\Omega\), a kernel function can be expanded as a unified weighted combination of basis functions, which is represented as
\begin{equation}
\label{universe}
\mathcal{K}(\delta )=\int_{\Omega }H(u)G(\delta ,u )\mathrm{d}u,
\end{equation}
where \( H(\cdot) \) represents the non-negative weights and \( G(\cdot,\cdot) \) represents the basis functions.

When \( \delta = x - y \), the integration domain $\Omega$ is \( \mathbb{R}^d \), the weights \( H(\cdot) \) are taken as a non-negative measure \( m(\cdot) \), and the basis functions \( G(\delta,u) = \exp(i u^T \delta) \), Equation (\ref{universe}) degenerates into Equation (\ref{shift-inv}), which is the expansion form of shift-invariant kernels with Fourier bases.

Schoenberg's theorem \cite{schoenberg} characterizes positive definite functions on the unit sphere in a Hilbert space, stating that a dot-product kernel $\mathcal{K}(\langle x, y \rangle)=\mathcal{K}(x,y):\ell_2 (0,1)\times \ell_2 (0,1)\to \mathbb{R}$ is positive definite if and only if \( \mathcal{K}(\cdot) \) is an analytic function admitting a Maclaurin expansion with only non-negative coefficients, written as
\begin{equation*}
\mathcal{K}(\left \langle x,y \right \rangle )= { \sum_{i=0}^{\infty }a_i {\left \langle x,y \right \rangle}^i},a_i\ge 0.
\end{equation*}

We note that if a discrete function \( f(z) \) is defined on $\mathbb{Z}$, such as the \( n \)-th term of the Maclaurin expansion of a dot-product kernel, its summation can be expressed as $\int_{\mathbb{R}}f(\left \lfloor u \right \rfloor )\mathrm{d}u=\sum_{z=0}^{\infty }f(z).$ Thus, under the guarantee of Schoenberg's theorem, let the integration domain $\Omega = \mathbb{Z}$, the weights \( H(\cdot) \) be the non-negative Maclaurin coefficients \( H(u)=\frac{1}{u!} \mathcal{K}^{(u)}(0) \), and the basis function \( G(\delta,u)=\delta ^u \) be polynomial basis, we obtain another instance of Equation (\ref{universe}) as
\begin{equation}
\label{polyexpand}
\mathcal{K}(\left \langle x,y \right \rangle ) =\int_{\mathbb{Z}}\frac{1}{u!} \mathcal{K}^{(u)}(0)\left \langle x,y \right \rangle^u \mathrm{d}u,
\end{equation}
where $\mathcal{K}^{(u)}(x)$ is the \( u \)-th derivative of \( \mathcal{K}(\cdot) \) at point \( x \). Schoenberg's theorem guarantees the equivalence between analytic functions with non-negative Maclaurin coefficients and dot-product kernels, where the coefficients represent the weights of the polynomial bases.

\begin{figure*}[t]
\centering
  \subfloat[Kernelized Attention]
  {
      \label{attnCal}\includegraphics[width=0.48\textwidth]{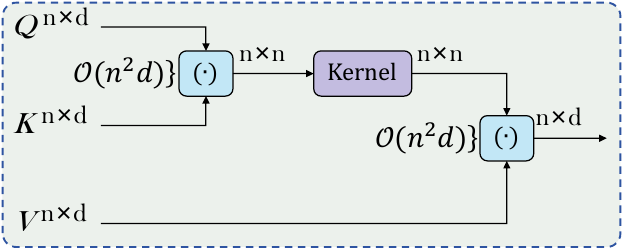}
  }
  \subfloat[RMFA]
  {
      \label{rmfaCal}\includegraphics[width=0.48\textwidth]{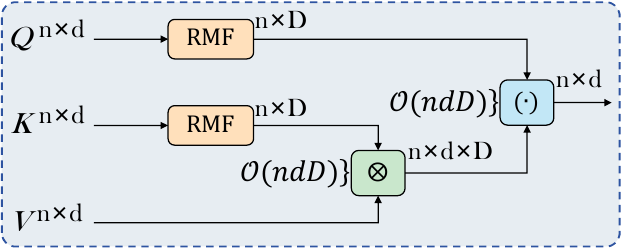}
  }
\caption{Computation graphs for kernelized attention and RMFA. In each figure, the data on the left represents the input to the attention layer. Here, operators $(\cdot)$ and $\otimes$ respectively denote matrix multiplication and outer product. Each step's main computation cost is marked on the left side of the operators, and the dimensions of data and intermediate results are indicated with superscripts.}
\label{calPro}
\end{figure*}

\subsection{Module Design of SchoenbAt}

We observe that kernel function \( \mathcal{K}(\bm{Q}\bm{K}^\top/\sqrt{d}) \) in kernelized attention is inherently a dot-product kernel. Based on Equation (\ref{polyexpand}), we expand dot-product kernelized attention and express it in the following form:
\begin{equation*}
{\rm attn}_\mathcal{K}(\bm{Q},\bm{K},\bm{V})\!=\!{ \sum_{i=1}^{n}\frac{\int_{\mathbb{Z}}\frac{1}{u!} \mathcal{K}^{(u)}(0)(\bm{Q}\bm{K}^\top/\sqrt{d})^u \mathrm{d}u\ \bm{V}_i }{ \sum_{j=1}^{n}\int_{\mathbb{Z}}\frac{1}{u!} \mathcal{K}^{(u)}(0)(\bm{Q}\bm{K}^\top/\sqrt{d})^u \mathrm{d}u \mathbf{1}_d^\top } }.
\end{equation*}

The polynomial expansion of \( \mathcal{K}(\left \langle x,y \right \rangle ) \) can be approximated by the inner product of RMF \cite{rmf} vectors \( \Phi_\mathcal{K}(x) \) and \( \Phi_\mathcal{K}(y) \) in arbitrary $D$s:
\begin{equation}
\label{rmfeq}
\mathbb{E}\left [\Phi_{\mathcal{K}}(x)\Phi^\top_{\mathcal{K}}(y)\right ]=\int_{\mathbb{Z}}\frac{1}{u!} \mathcal{K}^{(u)}(0)\left \langle x,y \right \rangle^u \mathrm{d}u,
\end{equation}
where \( \Phi_\mathcal{K}(x)=\sqrt{1/D} \left [ \phi_1 (x),\dots,\phi_D (x) \right ]\), each \( \phi_i (x)=\sqrt{a_N p^{N+1}}  {\textstyle \prod_{j=1}^{N} \left < \omega_j,x\right > } \) is independent of \( i \) , $p>1$ is a predetermined constant, \( N \) is sampled from \( \mathbb{N} \) with $\mathbb{P}[N=\eta]=\frac{1}{p^{\eta+1}} $, \( a_N \) is the \( N \)th coefficient of the Maclaurin expansion of the kernel function \( \mathcal{K}(\cdot) \), and \( \omega_j \in \mathbb{R}^d\) is a Rademacher vector independent of \( j \).

We establish the relationship between dot-product kernelized attention and the inner product of RMF through the polynomial basis expansion guaranteed by Schoenberg's theorem, thereby designing RMFA, a core component of SchoenbAt.

\begin{theorem}
\label{rmfath}
\textit{Suppose $\mathcal{K}(\bm{Q}\bm{K}^\top/\sqrt{d})$ is a dot-product kernel $\mathcal{K}(\bm{Q}/d^\frac{1}{4},\bm{K}/d^\frac{1}{4})$ with only non-negative Maclaurin coefficients, $ \Phi_\mathcal{K} (\cdot)$ defines a random Maclaurin feature map of $\mathcal{K}$, the output of the attention can be estimated by RMFA.} 
\begin{equation*}
\label{rmfaeq}
{\rm attn}_\mathcal{K}(\bm{Q},\bm{K},\bm{V})\approx\frac{ \Phi_\mathcal{K} (\bm{Q}/d^{\frac{1}{4}})  {\textstyle \sum_{i=1}^{n}\left [ \Phi_\mathcal{K}^\top (\bm{K}_i/d^{\frac{1}{4}})\right ]\otimes \bm{V}_i}}{ \Phi_\mathcal{K} (\bm{Q}/d^{\frac{1}{4}}) \sum_{j=1}^{n} \left [ \Phi_\mathcal{K} ^\top(\bm{K}_j/d^{\frac{1}{4}}) \right ] \mathbf{1}_d^\top}={\rm RMFA}_\mathcal{K}(\bm{Q},\bm{K},\bm{V}),
\end{equation*}
\end{theorem}
\noindent \textit{where $x \otimes y \in \mathbb{R}^{d\times d}$ represents the outer product of vectors.}

The proof is given in Appendix \ref{apdx:rmfa}. Figure \ref{calPro} illustrates the attention computation paths for the original kernelized attention and RFMA which factors out $\Phi_\mathcal{K}(\bm{Q}/d^{\frac{1}{4}}) $. In Figure \ref{attnCal}, the direct computation of the attention matrix results in a time complexity of \(O(n^2d)\). In contrast, the computation graph shown in Figure \ref{rmfaCal} changes the time complexity to \(O(ndD)\), which can be further decreased by parallelizing the outer product. In most cases, where \(n \gg D\), RMFA in SchoenbAt can significantly reduce the computational overhead of attention. In this paper, we study five dot product kernels with only non-negative Maclaurin coefficients as examples of SchoenbAt, which are shown in Table \ref{kernels}, where kernel \( {\exp}(\cdot) \) corresponds to Softmax attention. It is worth noting that the domains of dot-product kernels \({\rm inv}(\cdot) \), \({\rm logi}(\cdot) \), and \( {\rm sqrt}(\cdot) \) are restricted to inputs less than $1$.

\begin{figure}[t]
  \centering
  \begin{minipage}{0.54\textwidth}
    \centering
    \begin{tabular}{ccc}
      \toprule
      $\mathcal{K}$     & $f(\langle x,y \rangle )$     & $a_N$ \\
      \midrule
      $\mathrm{exp}$   & $\exp(\langle x,y \rangle )$    & $1/N!$ \\
      $\mathrm{inv}$   & $1/(1-\langle x,y \rangle )$    & $1$ \\
      $\mathrm{logi}$ & $1-\log(1-\langle x,y \rangle )$ & $\frac{1}{\min(1,N)}$ \\
      $\mathrm{trigh}$ & $\sinh(\langle x,y \rangle )+\cosh(\langle x,y \rangle )$ & $1/N!$ \\
      $\mathrm{sqrt}$  & $2-\sqrt{1-\langle x,y \rangle }$ & $\frac{\max(1,2N-3)}{2^N N!}$ \\
      \bottomrule
    \end{tabular}
    \captionof{table}{The dot-product kernels we study in this work along with their non-negative Maclaurin coefficients, where $\mathcal{K}(\cdot)=f(\langle x,y \rangle )$.}
    \label{kernels}
  \end{minipage}
  \hfill 
  \begin{minipage}{0.42\textwidth}
    \begin{algorithm}[H]
      \caption{ppSBN in SchoenbAt}
      \label{ppSBN}
      \textbf{Input}: attention input $\bm{Q}$, $\bm{K}$, $\bm{V}$ \\
      \textbf{Parameter}: trainable parameter $\beta$, $\gamma$; hyperparameter $\varepsilon$ \\
      \textbf{Output}: attention output $att$
      
      \begin{algorithmic}[1]
        \State $\bm{Q}',\bm{K}'\gets \frac{\bm{Q}-\bm{\mu_Q}}{\sqrt{\bm{\sigma_Q}+\bm{\varepsilon}}},\frac{\bm{K}-\bm{\mu_K}}{\sqrt{\bm{\sigma_K}+\bm{\varepsilon}}}$
        \State $\bm{Q}^{\mathrm{SBN}},\bm{K}^{\mathrm{SBN}}\gets \frac{\bm{Q}'}{\left\| \bm{Q}' \right\|_2 },\frac{\bm{K}'}{\left\| \bm{K}' \right\|_2 }$
        \State $att\gets \mathrm{RMFA}_\mathcal{K}(\bm{Q}^{\mathrm{SBN}},\bm{K}^{\mathrm{SBN}},\bm{V})$
        \State $att\gets {\gamma \cdot att}^\beta$
      \end{algorithmic}
    \end{algorithm}
  \end{minipage}
\end{figure}

According to Schoenberg's theorem, when the input space of kernelized attention is constrained to $\ell_2 (0,1)$, RMFA serves as an unbiased estimate of the dot-product kernelized attention. However, in practical applications, the input space of the attention module is uncontrolled, and the output will be strongly affected when the inputs are scaled. Inspired by the idea of reshaping data expectations and variances through Batch Normalization \cite{batchNorm}, SchoenbAt addresses this issue through a two-stage regularization mechanism called pre-post Scaling Batch Normalization.

As shown in Algorithm \ref{ppSBN}, SchoenbAt first normalizes and scales $\bm{Q},\bm{K}$ to ensure that $\bm{Q}^{\rm SBN},\bm{K}^{\rm SBN}\in \ell_2 (0,1)$, then feeds them into RMFA for computation. The mean value matrices $\bm{\mu_Q},\bm{\mu_K}$ and variance matrices $\bm{\sigma_Q},\bm{\sigma_K}$ are unsqueezed from vectors, and $\bm{\varepsilon}$ is the all-$\varepsilon$s matrix. The output of RMFA is rescaled by trainable parameters \( \beta \) and \( \gamma \), serving as the final output of SchoenbAt. We can mathematically demonstrate that when approximating \( {\rm attn}_{\rm exp} \), the ppSBN mechanism can restore the original output while ensuring that the input space remains constrained.

\begin{theorem}
\label{ppTheo}
\textit{Suppose $\bm{Q}^{\rm SBN},\bm{K}^{\rm SBN}$ denote $\bm{Q},\bm{K}$ after scaling and batch normalization, we have}
\begin{equation}
\label{ppsbn}
{\rm RMFA}_{\rm exp}(\bm{Q}^{\rm SBN},\bm{K}^{\rm SBN},\bm{V}) \approx{\frac{1}{t} {\left [ \frac{1}{s}\odot{\rm RMFA}_{\rm exp}(\bm{Q},\bm{K},\bm{V}) \right ] }^\frac{1}{r}} ,
\nonumber
\end{equation}

\noindent where $\odot$ denotes element-wise product, $r=\left \| \bm{Q}' \right \|_2 \left \| \bm{K}' \right \|_2 \sqrt{(\bm{\sigma_Q}+\bm{\varepsilon})(\bm{\sigma_K}+\bm{\varepsilon})}$ dependents on the data, $t=\sum_{i=1}^{n}{\rm exp}\left [ \left ( {\bm{Q}\bm{K}_i^\top-\bm{\mu_Q}\bm{K_i}^\top}\right )/{r\sqrt{d}}  \right ] \mathbf{1}_n^\top \big/\sum_{i=1}^{n}{\rm exp}\left [ \left ( {\bm{Q} \bm{K}_i^\top\mathbf{1}_n^\top-\bm{\mu_Q}\bm{K}^\top}\right )/{r\sqrt{d}} \right ]$ and $s=\frac{\left \| {\rm exp}(\bm{Q} \bm{K}^\top/\sqrt{d}) \right \|_{\frac{1}{r}} }{\left \| {\rm exp}(\bm{Q} \bm{K}^\top/\sqrt{d}) \right \|_1 }V^{r-1}$.
\end{theorem}

We give the proof of Theorem \ref{ppTheo} in Appendix \ref{apdx:ppsbn}. The trainable parameters \( \gamma \) and \( \beta \) in Algorithm \ref{ppSBN} fit the \( s\odot t^r \) and \( r \) in Theorem \ref{ppTheo}, thus preserving the scale of data. In addition, combining Theorem \ref{rmfath} and \ref{ppTheo}, we can obtain that the approximate equivalence is in the sense of expectation, i.e.
\begin{equation}
\label{ppsbneq}
\mathbb{E}\left [{\rm RMFA}_{\rm exp}(\bm{Q}^{\rm SBN},\bm{K}^{\rm SBN},\bm{V})\right ] ={\mathbb{E}\frac{1}{t} {\left [ \frac{1}{s}\odot{\rm RMFA}_{\rm exp}(\bm{Q},\bm{K},\bm{V}) \right ] }^\frac{1}{r}}.
\end{equation}

\begin{figure*}[t]
\centering
  \subfloat[Loss]
  {
      \label{ppsbn:subfig1}\includegraphics[width=0.32\textwidth]{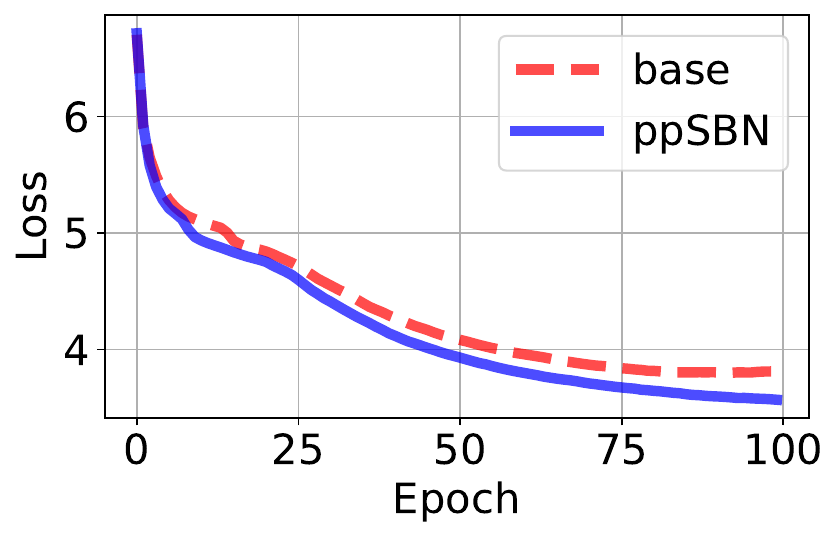}
  }
  \subfloat[PPL]
  {
      \label{ppsbn:subfig2}\includegraphics[width=0.32\textwidth]{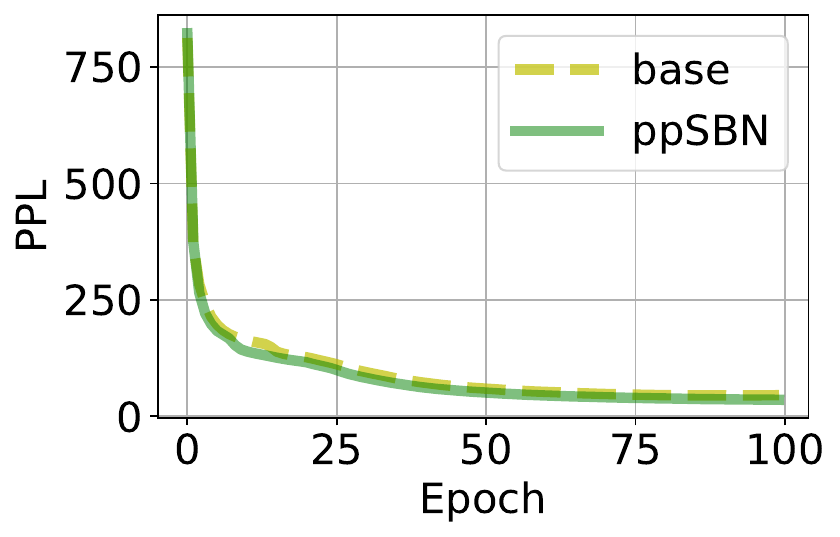}
  }
  \subfloat[BLEU]
  {
      \label{ppsbn:subfig3}\includegraphics[width=0.32\textwidth]{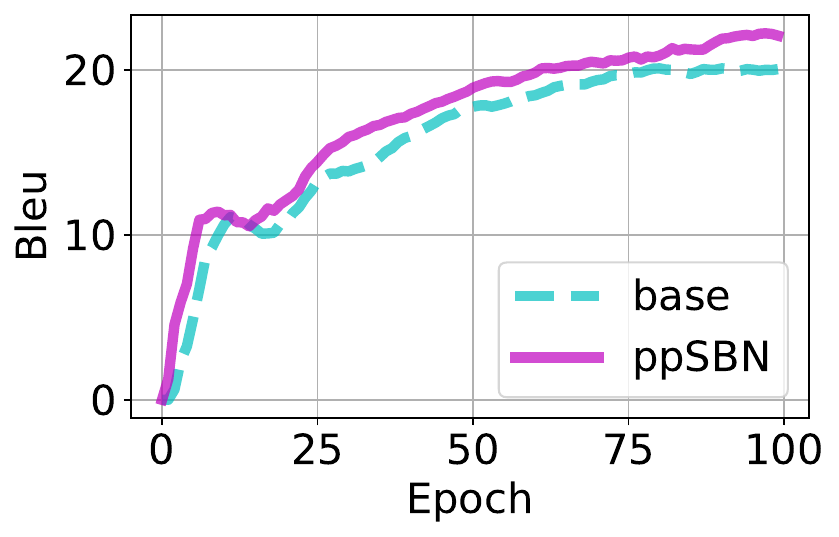}
  }
\caption{The loss, perplexity (lower is better), and Bleu scores (higher is better) of the traditional Transformer with and without ppSBN across training epochs on machine translation task of Multi30k dataset. In each plot, solid lines represent the Transformer with ppSBN, while dashed lines represent the Transformer without ppSBN.}
\label{ppsbnResult}
\end{figure*}

We conducted a toy experiment to verify whether the parameters \(\gamma\) and \(\beta\) in ppSBN can be appropriately trained. Figure \ref{ppsbnResult} shows the loss, perplexity, and Bleu scores of a machine translation experiment on the Multi30K dataset \cite{multi30k}, where we used the traditional Transformer \cite{transformer} as the base model to avoid the influence of randomness induced by RMFA and incorporated the ppSBN mechanism before and after the softmax attention layer for comparison. The model with the ppSBN mechanism slightly outperforms the base model in all three metrics, demonstrating that \(\gamma\) and \(\beta\) can be trained end-to-end with other parameters in a few epochs without degrading model performance.

Let $\overrightarrow{\rm SBN}(\bm{X}) :\bm{X}\to \bm{X}^{\rm SBN}$ represents the pre-SBN in algorithm \ref{ppSBN}, and $\overleftarrow{\rm SBN}(\bm{X}; \gamma,\beta) :\bm{X}\to (\gamma\bm{X})^\beta $ denotes the post-SBN, we provide SchoenbAt as
\begin{equation*}
{\rm SchoenbAt}_{\mathcal{K}}(\bm{Q},\bm{K},\bm{V})=\overleftarrow{\rm SBN}({\rm RMFA}_{\mathcal{K}}(\overrightarrow{\rm SBN}(\bm{Q}),\overrightarrow{\rm SBN}(\bm{K}),\bm{V}); \gamma,\beta).
\end{equation*}

As shown in Figure \ref{schnepic}, the input and output of SchoenbAt are the same as those of kernelized attention, making it a drop-in replacement for the attention mechanism in machine learning architectures.

\subsection{Theoretical Analyses of SchoenbAt}

Since the use of RMF introduces stochasticity into the approximation process, rigorous theoretical analysis is required to ensure the reliability and effectiveness of the proposed SchoenbAt mechanism. As an initial guarantee, we establish the unbiasedness formally proven in Appendix \ref{apdx:exp}.

\begin{theorem}
\label{expTheo}
\textit{Assuming that the parameters \(\gamma\) and \(\beta\) are properly trained to restore the attention output, $ \Phi_\mathcal{K} (\cdot):\mathbb{R}^d\to \mathbb{R}^D$ defines a random Maclaurin feature map for a dot-product kernel $\mathcal{K}(\cdot)$, then for any attention inputs $\bm{Q},\bm{K},\bm{V} \subset \mathbb{R}^{n \times d}$, we have $\mathbb{E}[{\rm SchoenbAt}_\mathcal{K}(\bm{Q},\bm{K},\bm{V})]={\rm attn}_\mathcal{K}(\bm{Q},\bm{K},\bm{V})$.}
\end{theorem}

Theorem \ref{expTheo} provides an important theoretical result, indicating that as the number of computations approaches infinity, the expectation of SchoenbAt converges to dot-product kernelized attention. This result establishes that, in the asymptotic case, SchoenbAt is an unbiased estimator of kernelized attention, offering a strong foundation for its approximation effectiveness. However, in practical learning tasks, SchoenbAt is computed only once during each forward pass, meaning that the approximation is subject to finite sampling and computational constraints, which impose strict requirements on estimation stability.

To address this issue, we derive an upper bound on the approximation error of SchoenbAt under the assumption that the norm of \( \bm{V} \) is bounded. This bound quantitatively characterizes the discrepancy between the output of SchoenbAt and that of the exact kernelized attention, thereby providing a theoretical measure of the approximation quality.

\begin{theorem}
\label{pacTheo}
\textit{Suppose attention inputs $\left | \bm{V}_{ij} \right | \le S$, assuming that the parameters \(\gamma\) and \(\beta\) are properly trained to restore the attention output, $ \Phi_\mathcal{K} (\cdot):\mathbb{R}^d\to \mathbb{R}^D$ defines a random Maclaurin feature map for a dot-product kernel $\mathcal{K}(\cdot)$. Let data $\mathcal{D}=(\bm{Q},\bm{K},\bm{V})$, then for any $\varepsilon >0$, we have}
\begin{equation*}
\begin{split}
\mathbb{P}\left ( \left | {\rm SchoenbAt}_\mathcal{K}(\mathcal{D}) \!-\!{\rm attn}_\mathcal{K}(\mathcal{D}) \right | > \epsilon \right )\le 2D\exp\!\left (- \frac{D\epsilon ^2}{2S^2d^2}  \right ) .
\end{split}
\end{equation*}
\end{theorem}

Detailed proof of Theorem \ref{pacTheo} is given in Appendix \ref{apdx:pac}. We have demonstrated that the approximation performance of SchoenbAt theoretically improves as \( D \) increases and \( S \) or \( d \) decreases. Specifically, increasing \( D \) leads to a more accurate approximation of kernelized attention. On the other hand, reducing \( S \) or \( d \) helps to minimize the scale of data which determines the cost of approximation.

In practical applications, the dimensionality of random features \( D \) can be adjusted flexibly to balance the approximation accuracy and computational speed. For example, a higher value of \( D \) may be suitable for scenarios where precision is critical, while a smaller \( D \) may be employed in real-time applications where speed is more important than perfect accuracy. This trade-off allows SchoenbAt to be efficiently deployed across a wide range of machine learning tasks with varying performance and resource requirements.

\begin{figure*}[tb]

\subfloat[Kernel $\exp(\cdot)$]
{
  \label{errExp}\includegraphics[width=0.19\textwidth]{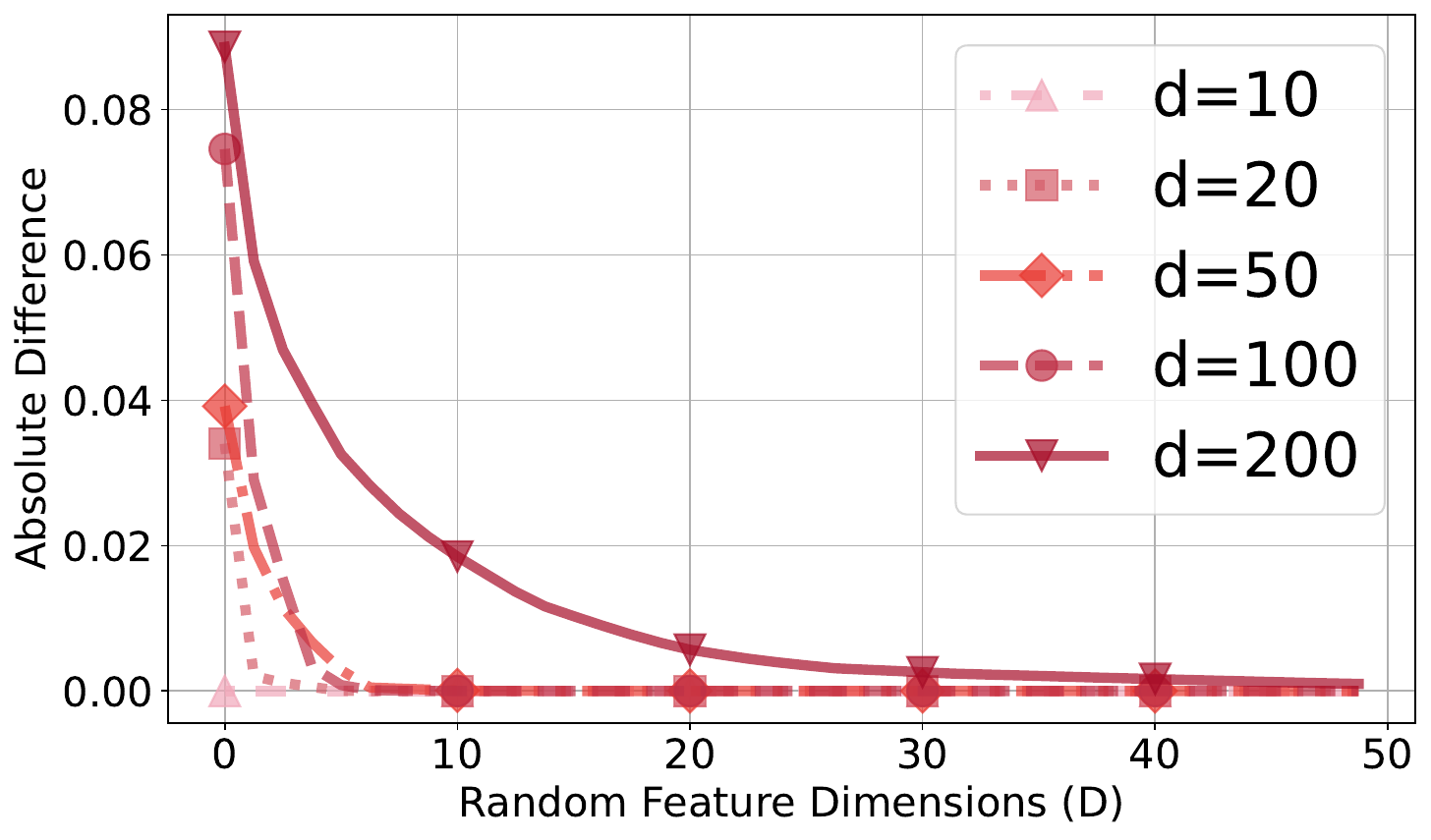}
}
\subfloat[Kernel {${\rm inv}(\cdot)$}]
{
  \label{errInv}\includegraphics[width=0.19\textwidth]{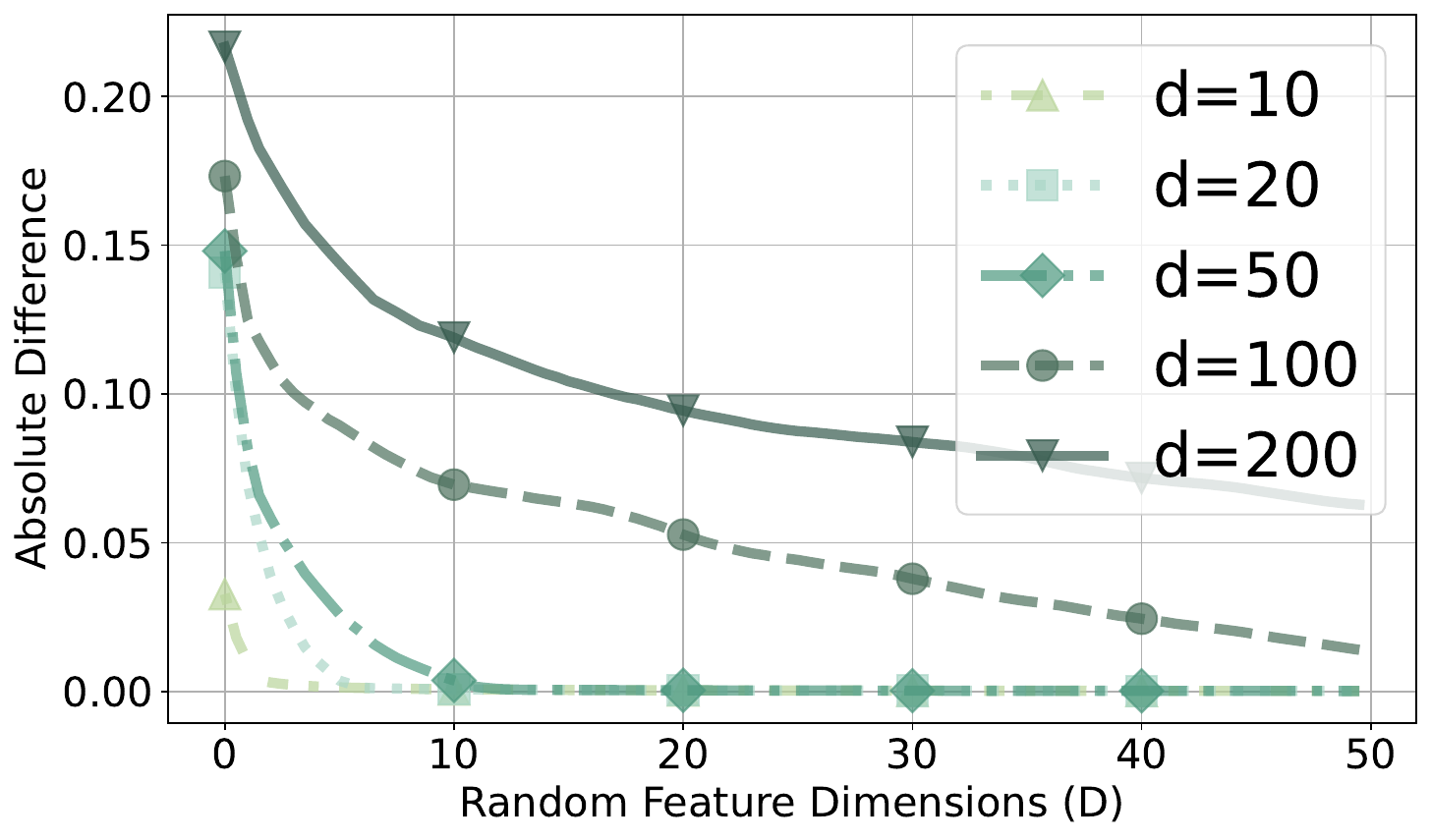}
}
\subfloat[Kernel ${\rm logi}(\cdot)$]
{
  \label{errLog}\includegraphics[width=0.19\textwidth]{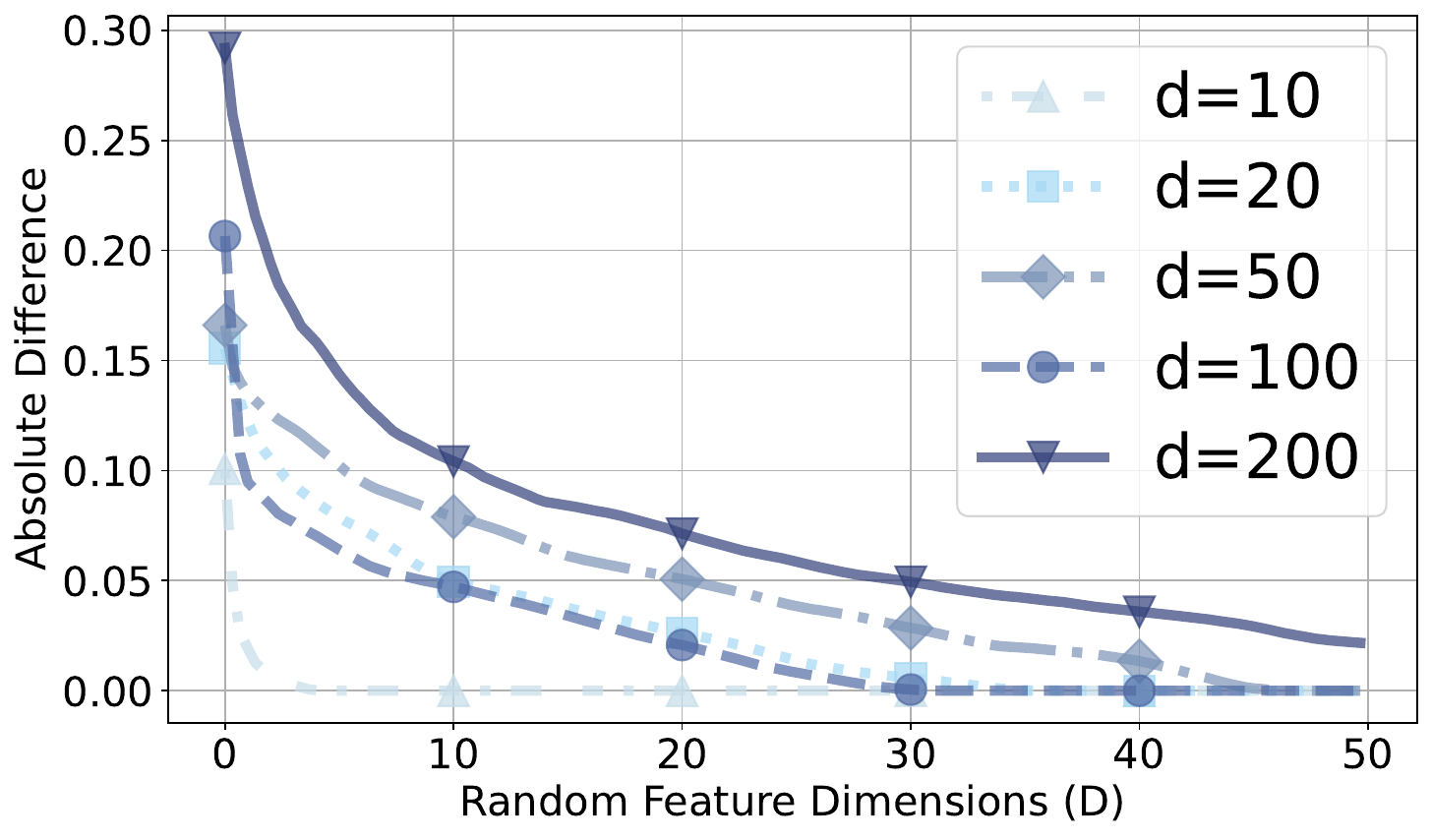}
}
\subfloat[Kernel ${\rm trigh}(\cdot)$]
{
  \label{errTrigh}\includegraphics[width=0.19\textwidth]{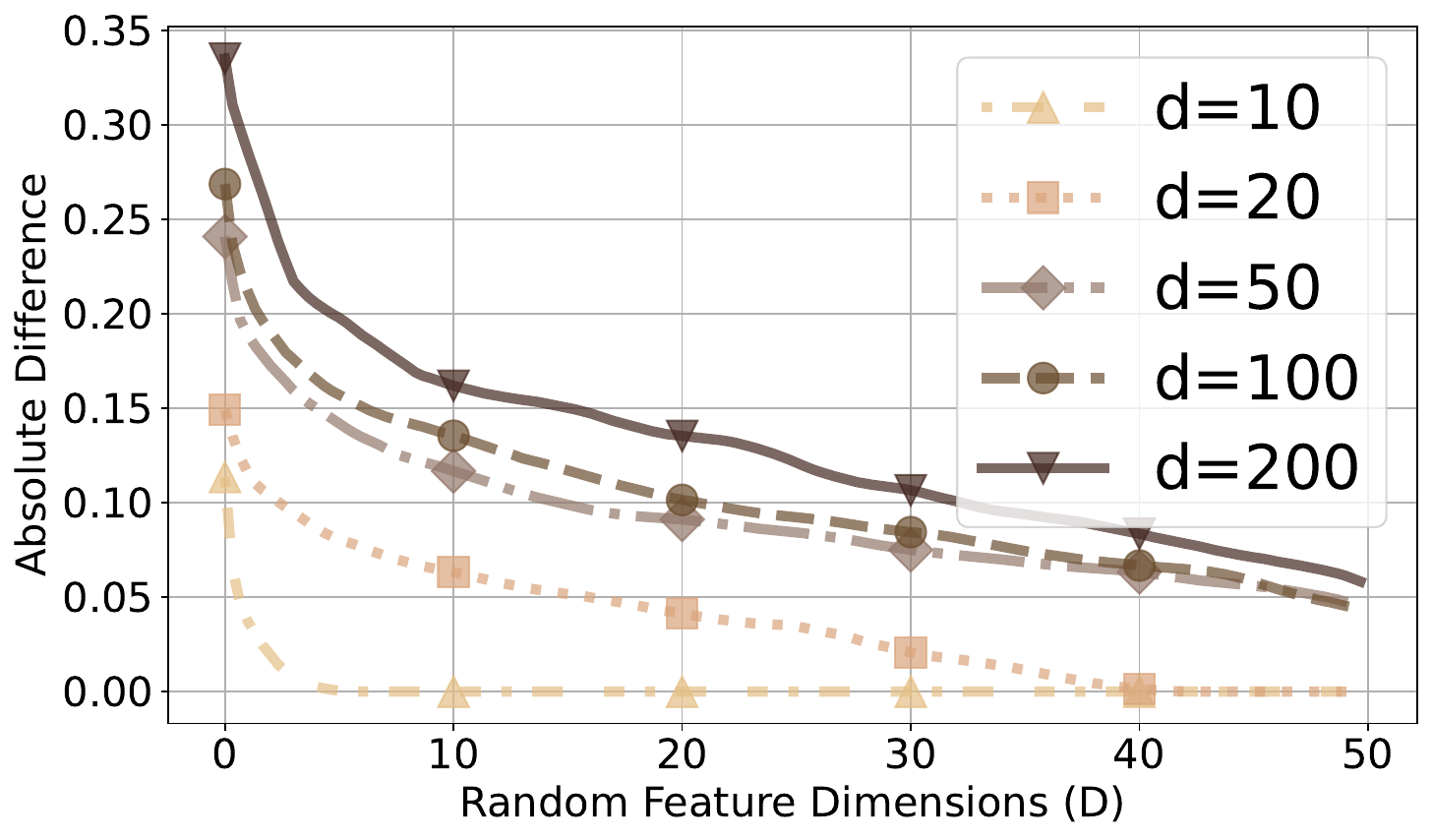}
}
\subfloat[Kernel ${\rm sqrt}(\cdot)$]
{
  \label{errSqrt}\includegraphics[width=0.19\textwidth]{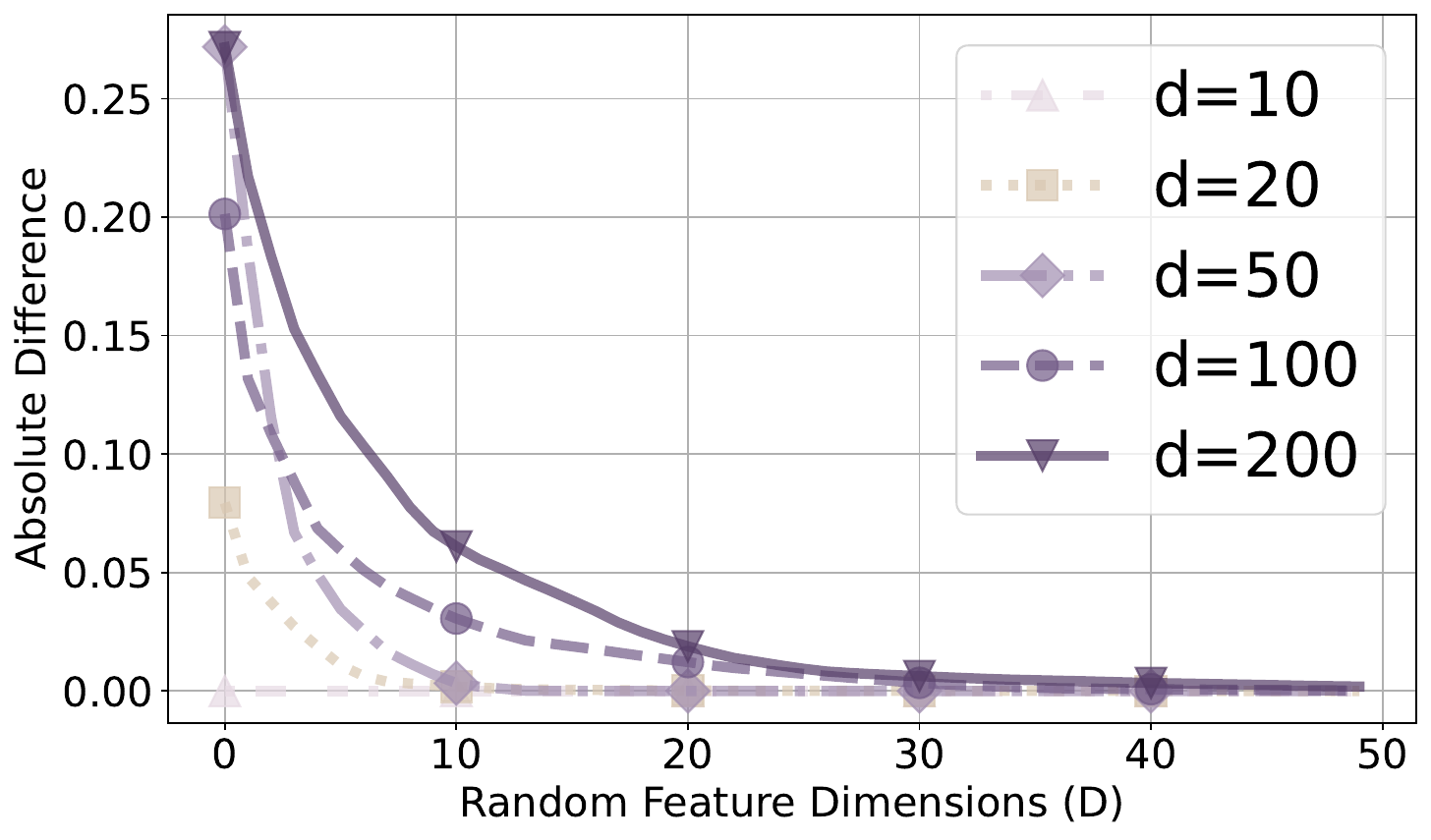}
}
\caption{The average absolute differences between SchoenbAt and kernelized attention on five dot-product kernels. Plots of different colors represent different dimensions of attention input.}
\label{erroring}
\end{figure*}

\begin{figure*}[tb]

\subfloat[Kernel ${\exp}(\cdot)$]
{
  \label{spdExp}\includegraphics[width=0.19\textwidth]{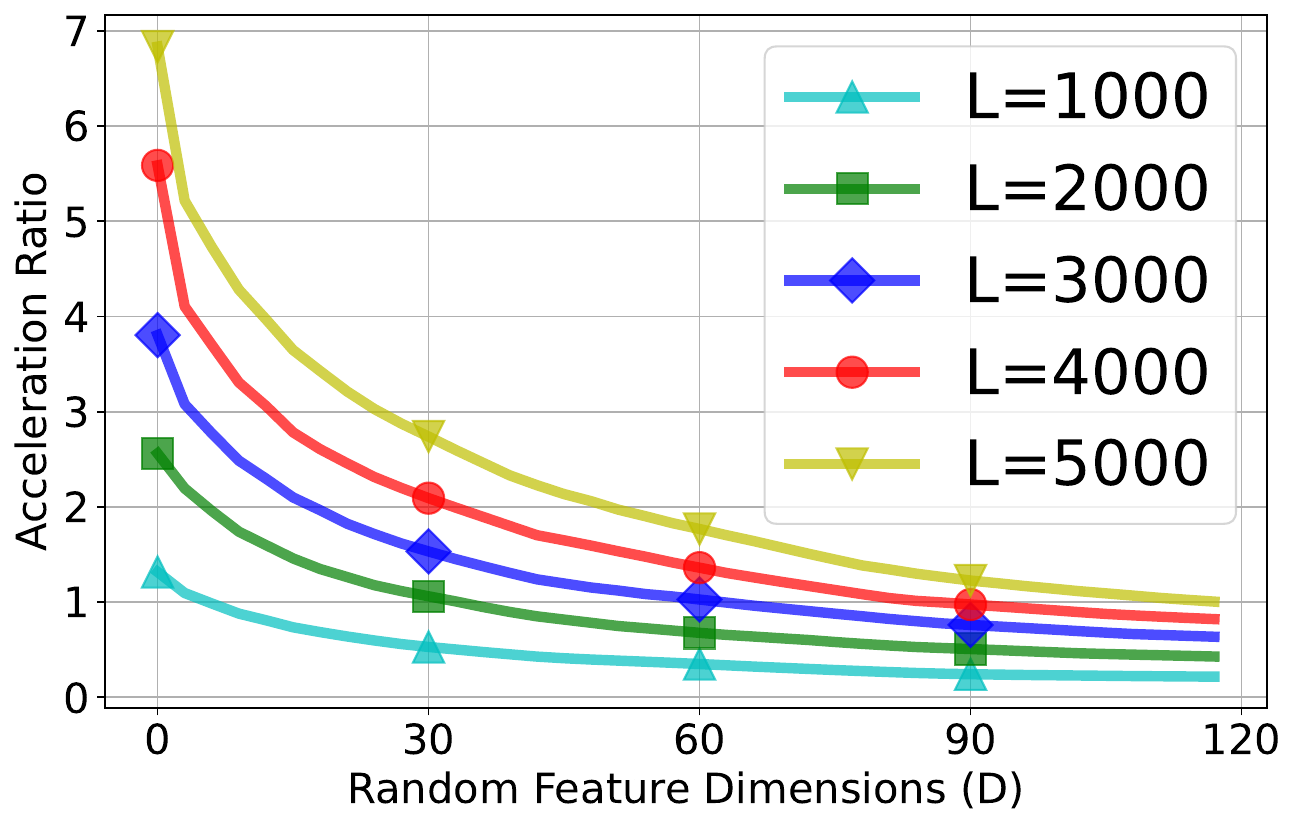}
}
\subfloat[Kernel {${\rm inv}(\cdot)$}]
{
  \label{spdInv}\includegraphics[width=0.19\textwidth]{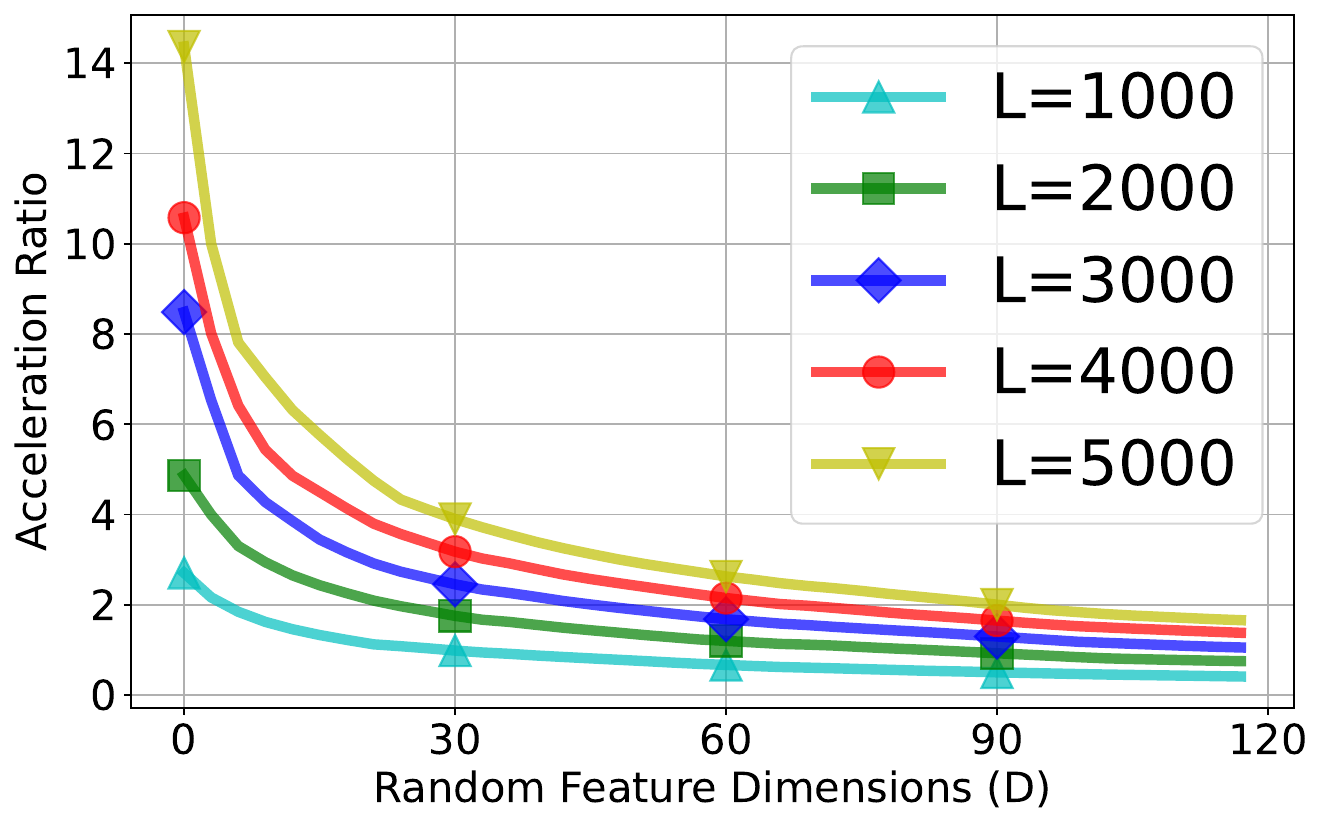}
}
\subfloat[Kernel ${\rm logi}(\cdot)$]
{
  \label{spdLog}\includegraphics[width=0.19\textwidth]{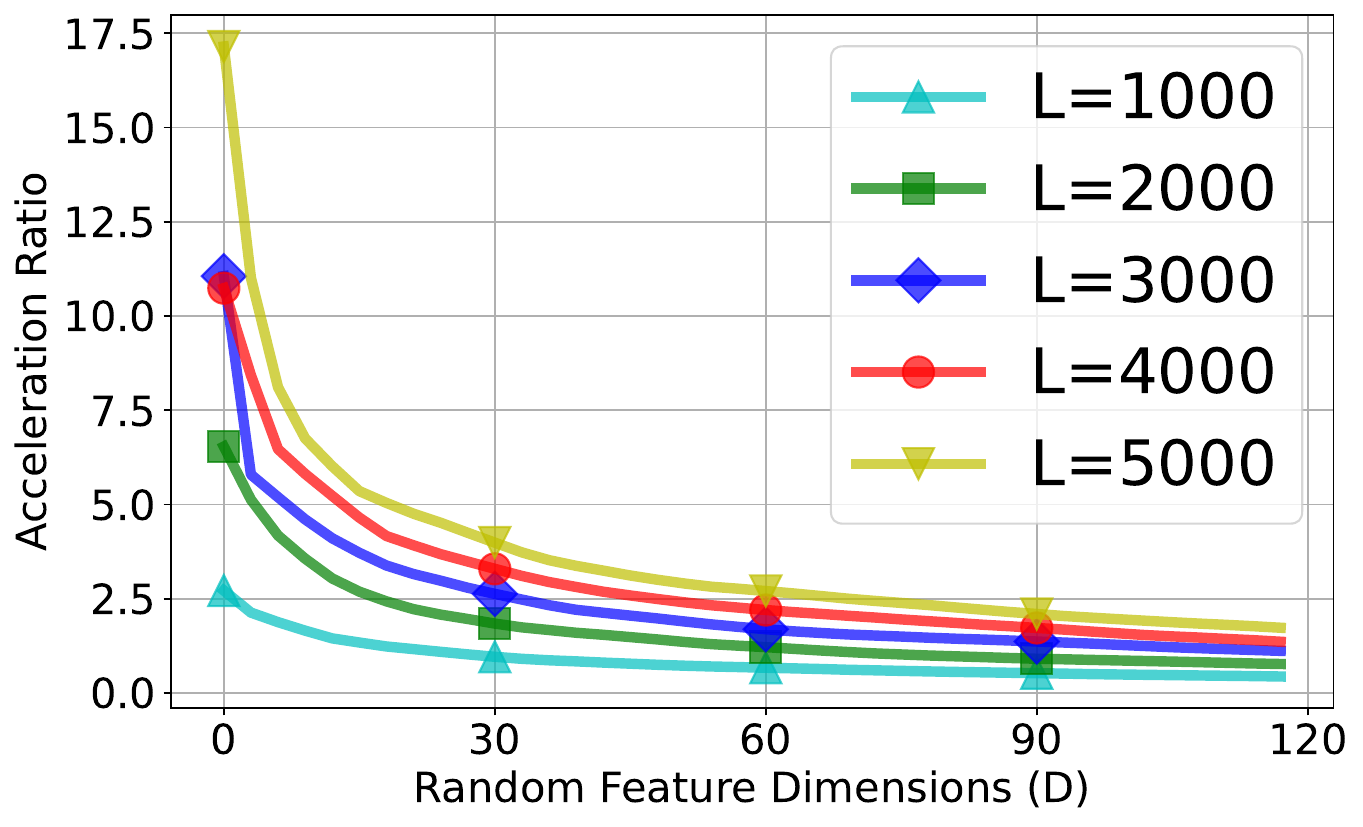}
}
\subfloat[Kernel ${\rm trigh}(\cdot)$]
{
  \label{spdTrigh}\includegraphics[width=0.19\textwidth]{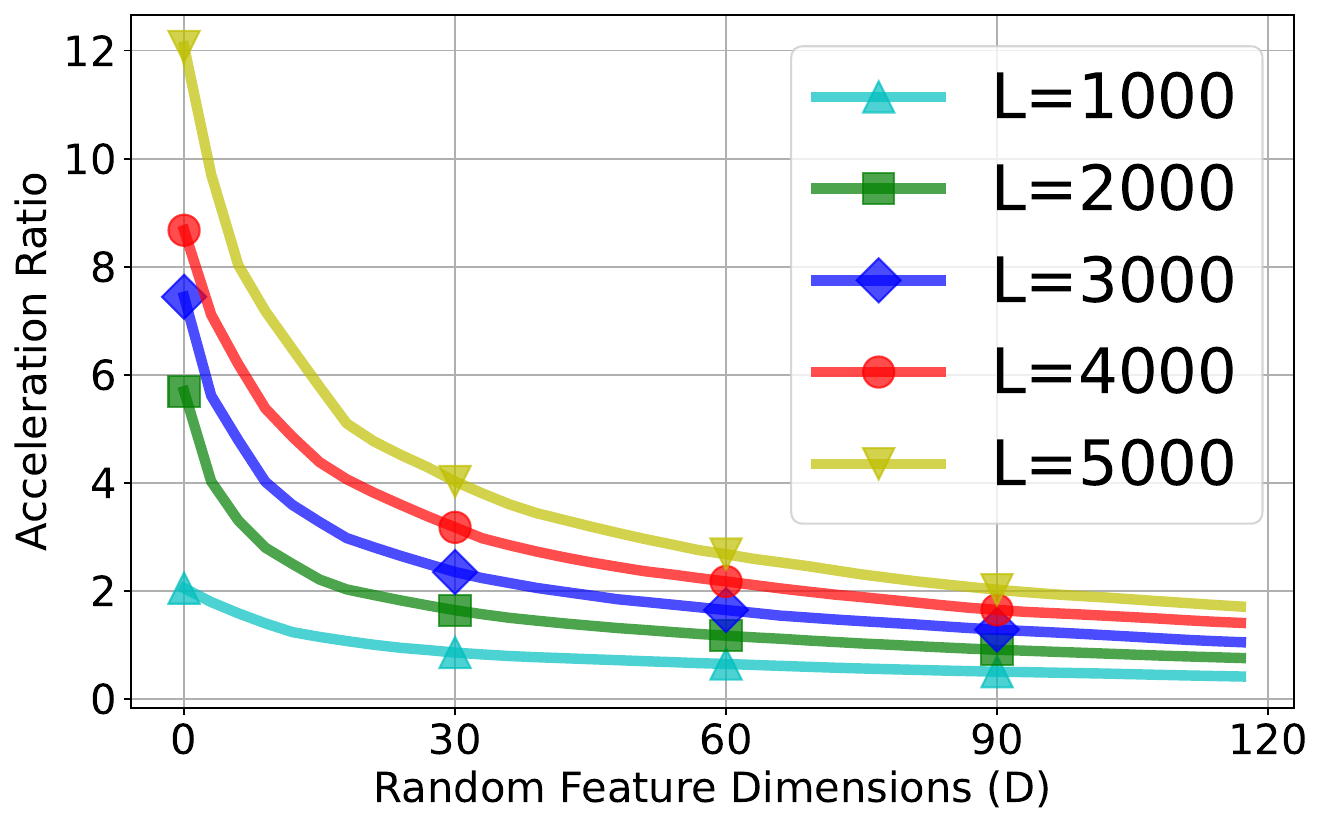}
}
\subfloat[Kernel ${\rm sqrt}(\cdot)$]
{
  \label{spdSqrt}\includegraphics[width=0.19\textwidth]{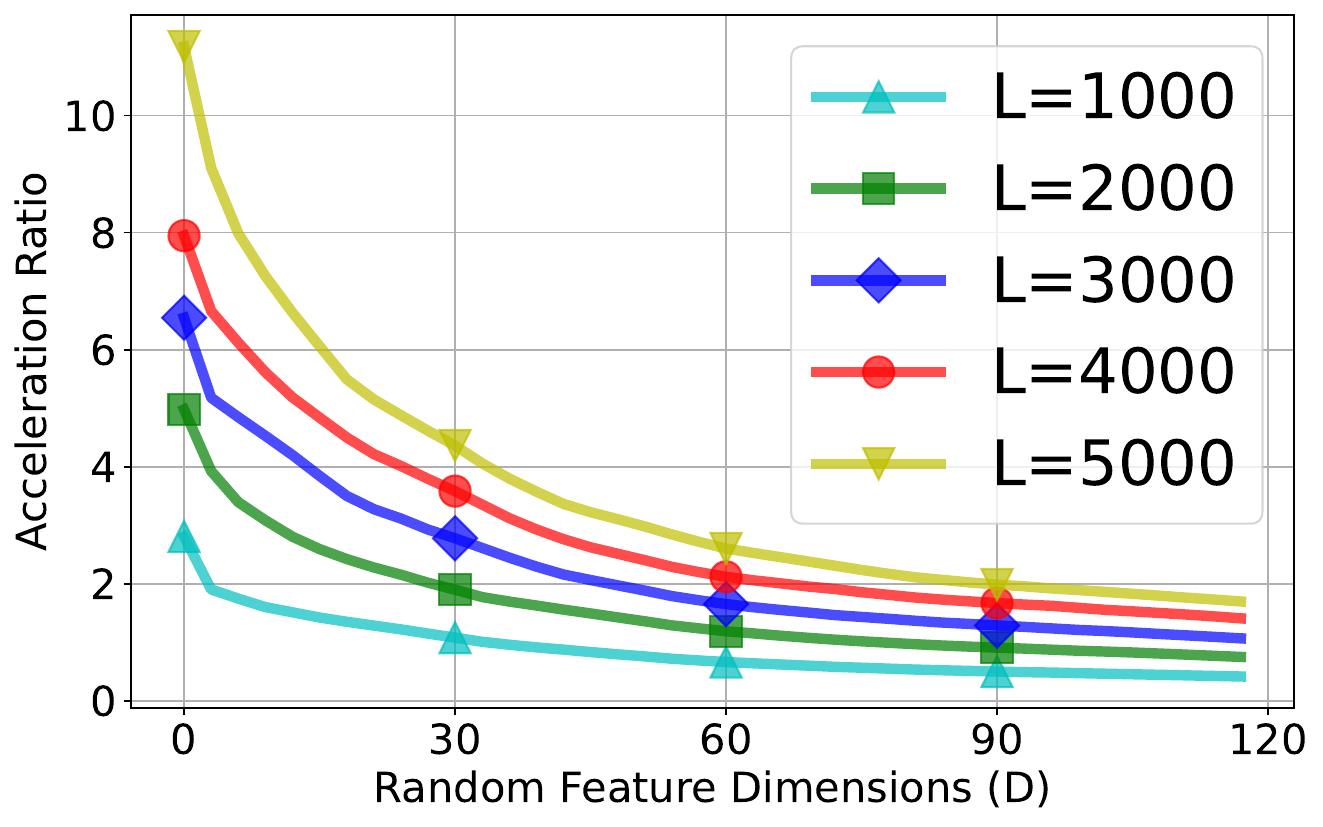}
}
\caption{The speedup ratio of SchoenbAt relative to kernelized attention on five dot-product kernels. Plots of different colors represent different sequence lengths of attention input.}
\label{speeding}
\end{figure*}

\section{Experiments}

In this section, we report a numerical experiment testing the speedup and error between SchoenbAt and kernelized attention, validations on the LRA benchmark, and an ablation study. We implement RMF based on Wacker \cite{randomFeature3} in each experiment and set the hyperparameter \( p = 2 \). Our code is available at \href{https://anonymous.4open.science/r/SchoenbAt-D6CE}{anonymous.4open.science/r/SchoenbAt-D6CE}.

\subsection{Approximation Error and Speedup}

In our first numerical experiment, we test the approximation error of SchoenbAt on five dot-product kernels in Table \ref{kernels}. We sample $\bm{Q}$, $\bm{K}$, and $\bm{V}$ from a \(100\times d\)-dimensional standard Gaussian distribution, where \(d\) varies from 10 to 200, and compute SchoenbAt with \(D\) varying from 10 to 50. The parameters \(\gamma\) and \(\beta\) are set to the ideally trained values. The experiment under each setting is repeated 100 times, and the error incurred by randomness is calculated as the average absolute difference between kernelized attention and SchoenbAt.

From Figure \ref{erroring}, one can easily observe that as \( D \) increases, the approximation error of SchoenbAt decreases rapidly, with the approximation error for ${\rm SchoenbAt}_{\rm \exp}$ being the smallest and decreasing the fastest, while the errors for ${\rm SchoenbAt}_{\rm logi}$ and ${\rm SchoenbAt}_{\rm trigh}$ are larger, with ${\rm SchoenbAt}_{\rm trigh}$ exhibiting the slowest rate of decrease. In each case, the increase in the original data dimension \( d \) leads to a higher approximation error for SchoenbAt, which still achieves a good approximation with a random feature dimension \( D \) that is lower than the data dimension \( d \). This experimental result confirms our Theorem \ref{pacTheo}.

Our second numerical experiment evaluates the speedup of SchoenbAt with five dot-product kernels relative to kernelized attention. The $\bm{Q}$, $\bm{K}$, and $\bm{V}$ were drawn from a standard Gaussian distribution with $d=50$, and the number of attention heads was set to 8. The sequence length $L$ varied between 1000 and 5000, while the random feature dimension \( D \) ranged from 2 to 120. Each experiment under the same configuration was repeated 100 times, and the speedup was calculated as the ratio of the time consumption of kernelized attention to that of SchoenbAt.

Figure \ref{speeding} illustrates a similar trend across all experiments: the speedup of SchoenbAt becomes more pronounced as \( L \) increases but diminishes as \( D \) grows. The speedup for ${\rm SchoenbAt}_{\rm logi}$ is notably higher than the others, while ${\rm SchoenbAt}_{\rm exp}$ shows the least speedup. When \( L \gg D \), SchoenbAt achieves remarkable acceleration, aligning with our complexity analysis.

\begin{table}[t]
\centering
\resizebox{\textwidth}{!}{
\begin{tabular}{cccccrrcccccccc}
\toprule
\multirow{2}[4]{*}{\textbf{Model}} & \multicolumn{1}{c}{\multirow{2}[4]{*}{\textbf{RF}}} & \multicolumn{6}{c}{\textbf{Time}}             &       & \multicolumn{6}{c}{\textbf{Accuracy (\%)}} \\
\cmidrule{3-8}\cmidrule{10-15}          &       & \textbf{T} & \textbf{L} & \textbf{R} & \multicolumn{1}{c}{\textbf{P}} & \multicolumn{1}{c}{\textbf{I}} & \textbf{Avg.} &       & \textbf{T} & \textbf{L} & \textbf{R} & \textbf{P} & \textbf{I} & \textbf{Avg.} \\
\midrule
Softmax & \ding{56}     & 1.000  & 1.000  & 1.000  & 1.000  & 1.000  & 1.000  &       & 63.31  & 37.30  & 75.04  & 68.64 & 38.69 & 56.59  \\
\midrule
Reformer & \ding{56}     & 0.181  & 0.229  & 0.330  & 0.851  & 0.484  & 0.415  &       & 64.76  & 23.99  & 73.36  & 50.29 & 37.23 & 49.92  \\
Nystromformer & \ding{56}     & 0.158  & \textcolor{red}{\textbf{0.188}}  & 0.280  & 0.484  & 0.482  & 0.318  &       & \textcolor{red}{\textbf{65.66}}  & 37.75  & 69.81  & 50.33 & 37.21 & 52.15  \\
Skyformer & \ding{56}     & 0.398  & 0.544  & 0.405  & 1.003  & 0.853  & 0.641  &       & 65.25  & 20.82  & 70.46  & 51.92 & 36.65 & 49.02  \\
Informer & \ding{56}     & 0.161  & 0.227  & 0.422  & 1.021  & 0.807  & 0.528  &       & 62.63  & 22.58  & 69.77  & 50.39 & 35.72 & 48.21  \\
Bigbird & \ding{56}     & 0.311  & 0.632  & 0.502  & 1.152  & 0.421  & 0.603  &       & 64.34  & 37.65  & 75.05  & 50.09 & 37.47 & 52.92  \\
Cosformer & \ding{56}     & 0.111  & 0.195  & 0.198  & \textcolor{red}{\textbf{0.265}}  & 0.271  & 0.208  &       & 63.84  & 23.49  & 75.32  & 50.27 & 33.72 & 49.32  \\
\midrule
RFA   & \ding{52}     & 0.783  & 0.836  & 0.900  & 0.727  & 0.866  & 0.822  &       & \textcolor{blue}{\textbf{65.22}}  & 37.15  & 77.84  & 59.91 & 36.09 & 55.24  \\
Performer & \ding{52}     & 0.349  & 0.412  & 0.294  & 0.623  & 0.455  & 0.426  &       & 63.79  & 18.80  & 75.69  & 50.37 & 31.98 & 48.12  \\
{${\rm SchoenbAt}_{\exp}$}   & \ding{52}     & \textcolor{red}{\textbf{0.076}}  & 0.236  & \textcolor{red}{\textbf{0.105}}  & 0.322  & \textcolor{red}{\textbf{0.214}}  & \textcolor{red}{\textbf{0.191}}  &       & 64.12  & 38.21  & 70.75  & \textcolor{red}{\textbf{72.35}} & 36.23 & 56.33  \\
{${\rm SchoenbAt}_{\rm inv}$}   & \ding{52}     & 0.209  & 0.193  & 0.194  & 0.412  & 0.268  & 0.255  &       & 63.54  & 38.56  & \textcolor{red}{\textbf{79.37}}  & 68.32 & 38.33 & 57.62  \\
{${\rm SchoenbAt}_{\rm logi}$}   & \ding{52}     & 0.167  & \textcolor{blue}{\textbf{0.192}}  & 0.106  & 0.437  & 0.289  & 0.238  &       & 64.62  & 39.77  & 70.35  & 71.42 & 38.35 & 56.90  \\
{${\rm SchoenbAt}_{\rm trigh}$} & \ding{52}     & 0.219  & 0.232  & 0.533  & 0.434  & 0.254  & 0.334  &       & 64.23  & \textcolor{red}{\textbf{39.92}}  & 70.71  & 71.39 & 38.35 & 56.92  \\
{${\rm SchoenbAt}_{\rm sqrt}$}  & \ding{52}     & 0.189  & 0.211  & 0.214  & \textcolor{blue}{\textbf{0.321}}  & 0.229  & 0.233  &       & 64.25  & 38.66  & 79.36  & 72.32 & \textcolor{red}{\textbf{38.49}} & \textcolor{red}{\textbf{58.61}}  \\
\bottomrule
\\
\end{tabular}%
}
\caption{Experimental results on three LRA benchmark tasks. We compared softmax attention, efficient attention improvements, and our method: SchoenbAt, where \textbf{RF} indicates whether it is a random feature method. We report the training time (normalized to Softmax) and accuracy on the test set, where \textbf{T}, \textbf{L}, \textbf{R}, \textbf{P}, \textbf{I} 
respectively represent Text, Listops, Retrieval, Pathfinder, Image tasks, and \textbf{Avg.} is the average across the three tasks. The data in \textcolor{red}{\textbf{red}} and \textcolor{blue}{\textbf{blue}} respectively represent the best performance among all methods and random feature methods.}
\label{macformerTable}%
\end{table}%
\subsection{Validations on LRA Benchmark}
\label{expriLRA}
In our experiments on real-world datasets, we replaced the attention mechanism in Transformer with SchoenbAt, Nyström-based attention \cite{reformer,Nyströmformer,skyformer}, linear attention \cite{informer,bigbird,cosformer}, and random 
feature-based attention \cite{rfa,performer}, comparing their performance on the LRA benchmark \cite{lra}. The LRA is a standard test used to assess and compare different Transformer models' ability to handle long sequence data. We organized our experiments around five tasks from this benchmark, consisting of natural language processing tasks: byte-level \textbf{\underline{T}ext} classification, long \textbf{\underline{L}istops}, byte-level document \textbf{\underline{R}etrieval}, and computer vision tasks: long-range \textbf{\underline{P}athfinder}, pixel-level \textbf{\underline{I}mage} classification.

We conducted all experiments on one NVIDIA RTX A6000 48G, working on a baseline model implemented by Chen \cite{skyformer} with PyTorch. Our model has an embedding dimension of $64$, a hidden dimension of $128$, $2$ layers, and uses $2$ attention heads. The batch size is selected depending on the task: $16$ for Text and Retrieval, $32$ for Listops, $128$ for Pathfinder and $256$ for Image. The default random projection dimension of each random feature-based method is set to $128$, and the $\epsilon$ for ppSBN is set to $10^{-13}$. Each training session involves $1000$ steps of initialization and $10000$ steps of optimization. All methods in our experiments share the same hyperparameters of the Transformer, and we report the average time consumption and accuracy over $50$ repetitions of training. 

Table \ref{macformerTable} presents the experimental results on the LRA benchmark, and memory cost is reported in Appendix \ref{resource} Table \ref{memtab}. In all five LRA tasks, SchoenbAt with five different dot-product kernels significantly improved computational efficiency while maintaining competitive accuracy. Notably, it outperformed other methods by a large margin in both the ListOps and Pathfinder tasks in terms of accuracy. Although SchoenbAt achieved less acceleration than Cosformer on the Pathfinder task, it yielded a substantial accuracy improvement: from $50.27\%$ to $71.16\%$ on average across the five kernels.Although SchoenbAt does not achieve the global optimum, it performs comparably to the best method in terms of runtime on the ListOps task and accuracy on the Text task, while significantly outperforming it on the other metric. Overall, ${\rm SchoenbAt}_{\rm exp}$ achieved the highest average speedup, while ${\rm SchoenbAt}_{\rm sqrt}$ attained the best average accuracy.

The evaluation on the LRA benchmark demonstrates SchoenbAt's capability in modeling both natural language dependencies and pixel-wise relationships. In addition, the above experimental results reveal a phenomenon: for each task, SchoenbAt with different kernel functions exhibits significantly varying performance. We believe this is because SchoenbAt approximates kernelized attention modeling sequence correlations with different kernel functions. We recommend using ${\rm SchoenbAt}_{\rm exp}$ in most cases, but other dot-product kernels can be substituted depending on the application scenario.

\begin{table*}[tp]
\resizebox{\textwidth}{!}{
\begin{tabular}{ccccccccccccc}
\toprule
\multicolumn{1}{c}{\multirow{2}[3]{*}{Ablations}} & \multirow{2}[3]{*}{base} & \multicolumn{5}{c}{base+RMFA}         & \multirow{2}[3]{*}{base+ppSBN} & \multicolumn{5}{c}{base+RMFA+ppSBN} \\
\cmidrule{3-7}\cmidrule{9-13}    \multicolumn{1}{c}{} &       & ${\exp(\cdot)}$   & ${{\rm inv}(\cdot)}$ & ${{\rm logi}(\cdot)}$  & ${{\rm trigh}(\cdot)}$ & ${{\rm sqrt}(\cdot)}$  &       & ${\exp(\cdot)}$   & ${{\rm inv}(\cdot)}$ & ${{\rm logi}(\cdot)}$  & ${{\rm trigh}(\cdot)}$ & ${{\rm sqrt}(\cdot)}$ \\
\midrule
Time  & 1.000  & 0.264  & 0.267  & 0.264  & 0.264  & 0.265  & 0.785  & 0.230  & 0.232  & 0.230  & 0.242  & 0.233  \\
Accuracy (\%) & 65.58 & 63.04 & 62.47 & 63.11 & 63.42 & 63.47 & 65.1  & 71.54 & 70.53 & 71.25 & 71.6  & 71.65 \\
\bottomrule
\\
\end{tabular}%
}
\caption{Ablation study on LRA-Text, reporting normalized training time and prediction accuracy under different settings, illustrating the individual and combined effects of RMFA and ppSBN.}%
\label{abtab}%
\end{table*}%

\subsection{Ablation Study}

To evaluate the individual contributions of RMFA and ppSBN in a controlled setting, we chose the Text task from LRA for ablation as it is the most representative and general-purpose sequence modeling benchmark across the dataset. Each experimental result is obtained as the average from leave-one-out cross-validation. We use standard softmax attention as the base model, and consider the following four configurations:

\begin{itemize}[leftmargin=10pt]
\item Base: The original Transformer with softmax attention, serving as a baseline.
\item Base+RMFA: Replacing softmax attention with RMFA with dot-product kernels in Table \ref{kernels}, measuring the effect of kernel approximation alone.
\item Base+ppSBN: Inserts ppSBN around softmax attention to examine its standalone impact.
\item Base+RMFA+ppSBN (SchoenbAt): Our full method combining both RMFA with dot-product kernels in Table \ref{kernels} and ppSBN.
\end{itemize}

This setup allows us to assess how RMFA contributes to efficiency and how ppSBN enhances the accuracy and stability of the approximation.

We report computational speedup (relative to the Base model) and prediction accuracy in Table \ref{abtab} to comprehensively evaluate each configuration. Using RMFA alone accelerates training but leads to a drop in accuracy, validating the approximation property claimed in Theorem \ref{rmfath}. In contrast, applying ppSBN alone yields a marginal improvement in efficiency while largely preserving or even improving accuracy, consistent with Theorem \ref{ppTheo}. SchoenbAt combines the strengths of both RMFA and ppSBN, achieving substantial speedup along with superior accuracy.

\section{Conclusion}

Under the guidance of Schoenberg's theorem, we expand dot-product kernelized attention and propose SchoenbAt as an efficient approximation. We rigorously establish its unbiasedness and derive approximation guarantees, both of which are validated through numerical simulations and empirical results on the LRA benchmark.

This work introduces a harmonic analysis-based theoretical framework for understanding and constructing kernel expansions in attention mechanisms. Future research may build upon this foundation by exploring alternative basis expansions or developing generalized random feature sampling techniques to approximate the unified form of kernelized attention more effectively.

{
\small
\bibliography{nips25}
}
\newpage
\appendix
\section*{Appendix}
\section{Proofs}

\subsection{Proof of Theorem \ref{rmfath}}
\label{apdx:rmfa}
\begin{proof}
We begin with
\begin{equation*}
{\rm attn}_\mathcal{K}(\bm{Q},\bm{K},\bm{V})={ \sum_{i=1}^{n}\frac{\int_{\mathbb{Z}}\frac{1}{u!} \mathcal{K}^{(u)}(0)(\bm{Q}\bm{K}^\top/\sqrt{d})^u \mathrm{d}u\ \bm{V}_i }{ \sum_{j=1}^{n}\int_{\mathbb{Z}}\frac{1}{u!} \mathcal{K}^{(u)}(0)(\bm{Q}\bm{K}^\top/\sqrt{d})^u \mathrm{d}u \mathbf{1}_d^\top} },
\end{equation*}
by Equation (\ref{rmfeq}) we have $\int_{\mathbb{Z}}\frac{1}{u!} \mathcal{K}^{(u)}(0)(\bm{Q}\bm{K}^\top/\sqrt{d})^u \mathrm{d}u\approx \Phi_\mathcal{K}(\bm{Q}/d^{\frac{1}{4}}) \Phi_\mathcal{K}^\top(\bm{K}_i/d^{\frac{1}{4}})$, thus

\begin{align*}
\label{rmfaeq}
\begin{split}
&\quad\ {\rm attn}_\mathcal{K}(\bm{Q},\bm{K},\bm{V}) \approx{{ \sum_{i=1}^{n}\frac{[\Phi_\mathcal{K}(\bm{Q}/d^{\frac{1}{4}}) \Phi_\mathcal{K}^\top(\bm{K}_i/d^{\frac{1}{4}})] \bm{V}_i }{ \sum_{j=1}^{n}\Phi_\mathcal{K}(\bm{Q}/d^{\frac{1}{4}}) \Phi_\mathcal{K}^\top(\bm{K}_j/d^{\frac{1}{4}})\mathbf{1}_d^\top} }}\\
&=\frac{ \Phi_\mathcal{K} (\bm{Q}/d^{\frac{1}{4}})  {\textstyle \sum_{i=1}^{n}\left [ \Phi_\mathcal{K}^\top (\bm{K}_i/d^{\frac{1}{4}})\right ]\otimes \bm{V}_i}}{ \Phi_\mathcal{K} (\bm{Q}/d^{\frac{1}{4}}) \sum_{j=1}^{n} \left [ \Phi_\mathcal{K} ^\top(\bm{K}_j/d^{\frac{1}{4}}) \right ]\mathbf{1}_d^\top }={\rm RMFA}_\mathcal{K}(\bm{Q},\bm{K},\bm{V}).
\end{split}
\end{align*}
\end{proof}

\subsection{Proof of Theorem \ref{ppTheo}}
\label{apdx:ppsbn}
\begin{proof}
By Theorem \ref{rmfath} we have ${\rm RMFA}_{\rm exp}(\bm{Q}^{\rm SBN}\!,\bm{K}^{\rm SBN}\!,\bm{V})\approx{\rm attn}_{\rm exp}(\bm{Q}^{\rm SBN}\!,\bm{K}^{\rm SBN}\!,\bm{V})$, let $r=\left \| \bm{Q}' \right \|_2 \left \| \bm{K}' \right \|_2 \sqrt{(\bm{\sigma_Q}+\bm{\varepsilon})(\bm{\sigma_K}+\bm{\varepsilon})}$ depends on the data, $t=\frac{\sum_{i=1}^{n}{\rm exp}\left [ \left ( {\bm{Q}\bm{K}_i^\top-\bm{\mu_Q}\bm{K_i}^\top}\right )/{r\sqrt{d}}  \right ] \mathbf{1}_n^\top}{ \sum_{i=1}^{n}{\rm exp}\left [ \left ( {\bm{Q} \bm{K}_i^\top\mathbf{1}_n^\top-\bm{\mu_Q}\bm{K}^\top}\right )/{r\sqrt{d}} \right ] }$ and $s=\frac{\left \| {\rm exp}(\bm{Q} \bm{K}^\top/\sqrt{d}) \right \|_{\frac{1}{r}} }{\left \| {\rm exp}(\bm{Q} \bm{K}^\top/\sqrt{d}) \right \|_1 }V^{r-1}$, we have
\begin{equation}
\begin{split}
&\quad\ {\rm RMFA}_{\rm exp}(\bm{Q}^{\rm SBN},\bm{K}^{\rm SBN},\bm{V})\approx{\rm Softmax}\left [ \frac{(\bm{Q}-\bm{\mu_Q})(\bm{K}-\bm{\mu_K})^\top }{\left \| \bm{Q}' \right \|_2 \left \| \bm{K}' \right \|_2 \sqrt{(\bm{\sigma_Q}+\bm{\varepsilon} )(\bm{\sigma_K}+\bm{\varepsilon})}} \bigg/\sqrt d\right ] \bm{V}\\
&=\frac{\sum_{i=1}^{n}{\rm exp}\left [ \left ( {\bm{Q}\bm{K}_i^\top\mathbf{1}_n^\top-\bm{\mu_Q}\bm{K}^\top}\right )/{r\sqrt{d}}  \right ] }{ \sum_{i=1}^{n}{\rm exp}\left [ \left ( {\bm{Q} \bm{K}_i^\top-\bm{\mu_Q}\bm{K}_i^\top}\right )/{r\sqrt{d}} \right ]\mathbf{1}_n^\top } \frac{{\rm exp}(\bm{Q} \bm{K}^\top/{r\sqrt{d}})\bm{V}}{ {\textstyle \sum_{i=1}^{n}{\rm exp}(\bm{Q} \bm{K}_i^\top/{r\sqrt{d}})\mathbf{1}_d^\top} }\\
&=\frac{1}{t}\sqrt[r]{\frac{\left \| {\rm exp}(\bm{Q} \bm{K}^\top/\sqrt{d}) \right \|_1 }{\left \| {\rm exp}(\bm{Q} \bm{K}^\top/\sqrt{d}) \right \|_{\frac{1}{r}} }\bm{V}^{r-1}\odot \frac{{\rm exp}(\bm{Q} \bm{K}^\top/\sqrt{d})\bm{V}}{ {\textstyle \sum_{i=1}^{n}{\rm exp}(\bm{Q} \bm{K}_i^\top\mathbf{1}_d^\top/\sqrt{d})} }  }\\
&=\frac{1}{t}\sqrt[r]{ \frac{1}{s}\odot {\rm attn}_{\rm exp}(\bm{Q},\bm{K},\bm{V})}\approx{\frac{1}{t} {\left [ \frac{1}{s}\odot{\rm RMFA}_{\rm exp}(\bm{Q},\bm{K},\bm{V}) \right ] }^\frac{1}{r}},
\end{split}
\nonumber
\end{equation}

Combinign Theorem \ref{rmfath} and \ref{ppTheo}, we can get
\begin{equation}
\begin{split}
&\quad\ \mathbb{E}\left [ {\rm RMFA}_{\rm exp}(\bm{Q}^{\rm SBN},\bm{K}^{\rm SBN},\bm{V})\right ]=\mathbb{E}\left [ {\rm attn}_{\rm exp}(\bm{Q}^{\rm SBN}\!,\bm{K}^{\rm SBN}\!,\bm{V})\right ]\\
&=\mathbb{E}\left [ \frac{1}{t}\sqrt[r]{ \frac{1}{s}\odot {\rm attn}_{\rm exp}(\bm{Q},\bm{K},\bm{V})}\right ]=\mathbb{E}{\frac{1}{t} {\left [ \frac{1}{s}\odot{\rm RMFA}_{\rm exp}(\bm{Q},\bm{K},\bm{V}) \right ] }^\frac{1}{r}}.
\end{split}
\nonumber
\end{equation}

\end{proof}

\subsection{Proof of Theorem \ref{expTheo}}
\label{apdx:exp}
\begin{proof}
We begin with the expectation-wise equality of ppSBN, as given in Equation (\ref{ppsbneq}),
\begin{equation*}
\mathbb{E}[{\rm SchoenbAt}_\mathcal{K}(\bm{Q},\bm{K},\bm{V})]=\mathbb{E}\left [\gamma{\rm RMFA}_{\mathcal{K}}(\overrightarrow{\rm SBN}(\bm{Q}),\overrightarrow{\rm SBN}(\bm{K}),\bm{V})^\beta\right ]=\mathbb{E}\left [{\rm RMFA}_{\mathcal{K}}(\bm{Q},\bm{K},\bm{V})\right ].
\end{equation*}
With the fact that $d\ge 1$, we can ensure that $\bm{Q}/d^{\frac{1}{4}},\bm{K}/d^{\frac{1}{4}}\in \ell_2 (0,1)$, then we have
\begin{align*}
\nonumber
&\quad\  \mathbb{E}\left [{\rm RMFA}_{\mathcal{K}}(\bm{Q},\bm{K},\bm{V})\right ]=\sum_{i=1}^{n}\frac{\mathbb{E} \left [ \frac{1}{D} {\textstyle \sum\phi(\bm{Q}/d^{\frac{1}{4}})\phi^\top(\bm{K}_i/d^{\frac{1}{4}})} \right ] \bm{V}_i }{ \sum_{j=1}^{n}\mathbb{E} \left [ \frac{1}{D} {\textstyle \sum\phi(\bm{Q}/d^{\frac{1}{4}})\phi^\top(\bm{K}_j/d^{\frac{1}{4}})} \right ] \mathbf{1}_d^\top}\\
&=\sum_{i=1}^{n}\!\!\frac{\mathbb{E}\!\! \left [ \frac{1}{D} {\textstyle \sum\!\!\phi(\bm{Q}/d^{\frac{1}{4}})\phi^\top\!(\bm{K}_i/d^{\frac{1}{4}})} \right ] \bm{V}_i }{ \sum_{j=1}^{n}\!\mathbb{E}\!\! \left [ \frac{1}{D} {\textstyle \sum\!\!\phi(\bm{Q}/d^{\frac{1}{4}})\phi^\top\!\!(\bm{K}_j/d^{\frac{1}{4}})} \right ] \!\mathbf{1}_d^\top}\!\! =\!\!\sum_{i=1}^{n}\!\!\frac{\mathbb{E}_N\!\! \left [{a_N p^{N+1} \mathbb{E}_\omega\!\! \left ( \prod_{l=1}^N\!\! \left \langle \omega_l ,\bm{Q} \right \rangle\!\! \left \langle \omega_l ,\bm{K}_i \right \rangle\! /\!\sqrt d \right ) } \right ] \bm{V}_i }{ \sum_{j=1}^{n}\mathbb{E}_N\!\! \left [ {a_N p^{N+1} \mathbb{E}_\omega\!\! \left ( \prod_{l=1}^N\!\! \left \langle \omega_l ,\bm{Q}\right \rangle \!\!\left \langle \omega_l ,\bm{K}_j\right \rangle\!  /\!\sqrt d \right ) } \right ]\! \mathbf{1}_d^\top}\\
&=\sum_{i=1}^{n}\frac{ {\textstyle \sum_{\eta =0}^{\infty}a_\eta \frac{1}{p^{\eta+1}}p^{\eta+1}(\bm{Q} \bm{K}_i^\top/{\sqrt d})^\eta }\  \bm{V}_i }{ \sum_{j=1}^{n} {\left [ \textstyle \sum_{\eta =0}^{\infty}a_\eta \frac{1}{p^{\eta+1}}p^{\eta+1}(\bm{Q} \bm{K}_j^\top/{\sqrt d})^\eta \right ] } \mathbf{1}_d^\top}={ \sum_{i=1}^{n}\frac{\mathcal{K}(\bm{Q} \bm{K}_i^\top/\sqrt{d}) \bm{V}_i }{ \sum_{j=1}^{n}\mathcal{K}(\bm{Q} \bm{K}_j^\top/\sqrt{d}) \mathbf{1}_d^\top} }={\rm attn}_\mathcal{K}(\bm{Q},\bm{K},\bm{V}).
\end{align*}


\end{proof}

\subsection{Proof of Theorem \ref{pacTheo}}
\label{apdx:pac}

\begin{proof}
Since the randomness in SchoenbAt only affects the attention matrix, and the attention mechanism computes a linear combination of $\bm{V}$, we have
\begin{equation}
\forall \left | \bm{V}_{ij} \right | \le S,
{\rm SchoenbAt}_\mathcal{K}(\mathcal{D})_{ij}\in [-S d,S d ],
\nonumber
\end{equation}

then, by Hoeffding's inequality,
\begin{align*}
&\quad\ \mathbb{P}\left ( \left | {\rm SchoenbAt}_\mathcal{K}(\mathcal{D}) -{\rm attn}_\mathcal{K}(\mathcal{D}) \right | > \epsilon \right )=\mathbb{P}\left ( \left | {\rm RMFA}_\mathcal{K}(\mathcal{D}) -{\rm attn}_\mathcal{K}(\mathcal{D}) \right | > \epsilon \right )\\[6pt]
&=\mathbb{P}\left ( \exists\phi_t:\Bigg | \sum_{i=1}^{n}\frac{ \left [ {\phi_t(\bm{Q}/d^\frac{1}{4} )\phi_t(\bm{K}_i/d^\frac{1}{4})} \right ]  \bm{V}_i }{ \sum_{j=1}^{n} {\phi_t(\bm{Q}/d^\frac{1}{4})\phi_t(\bm{K}_j/d^\frac{1}{4})}\mathbf{1}_n^\top}-{\rm attn}_\mathcal{K}(\mathcal{D}) \Bigg | >\epsilon \right ) \\
&\le \sum_{t=1}^{D} {\mathbb{P}\Bigg ( \Bigg | \sum_{i=1}^{n}\frac{ \left [ {\phi_t(\bm{Q}/d^\frac{1}{4})\phi_t(\bm{K}_i/d^\frac{1}{4})} \right ]  \bm{V}_i }{ \sum_{j=1}^{n} {\phi_t(\bm{Q}/d^\frac{1}{4})\phi_t(\bm{K}_j/d^\frac{1}{4})} \mathbf{1}_n^\top}-{\rm attn}_\mathcal{K}(\mathcal{D}) \Bigg | >\epsilon \Bigg )}\\
&\le D\cdot 2{\rm exp}\left [ -\frac{2D^2\epsilon ^2}{D(2S d)^2}  \right ] =2D\ {\rm exp}\left (- \frac{D\epsilon ^2}{2S^2d^2} \right ).
\nonumber
\end{align*}
\end{proof}

\section{Experiment Compute Resources}
\label{resource}

\begin{table}[ht]
  \centering
  \resizebox{0.8\textwidth}{!}{
      \begin{tabular}{ccccccc}
    \toprule
    Model & \textbf{T} & \textbf{L} & \textbf{R} & \textbf{P} & \textbf{I} & \textbf{Avg.} \\
    \midrule
    Softmax & 18092 & 4878  & 11284 & 6058  & 11790 & 10420  \\
    SchoenbAt & 4452  & 1696  & 3758  & 2062  & 3636  & 3121  \\
    Ratio (\%) & 0.246  & 0.348  & 0.333  & 0.340  & 0.308  & 0.315  \\
    \bottomrule
    \\
    \end{tabular}%
  }
  \caption{Memory consumption of SchoenbAt compared to Softmax attention in Experiment \ref{expriLRA}. Since SchoenbAt with the five dot-product kernels consumes the same amount of memory, we represent them using a single row in the table.}
  \label{memtab}%
\end{table}%

\end{document}